\documentclass[preprint,authoryear,a4paper]{elsarticle}

\usepackage{verbatim}

\usepackage{geometry}

\usepackage{color,soul}

\usepackage[hidelinks]{hyperref}
\hypersetup{
  colorlinks   = true, 
  urlcolor     = blue,
  linkcolor    = blue, 
  citecolor    = blue 
}

\usepackage{graphicx}
\graphicspath{{./}}
\DeclareGraphicsExtensions{.pdf,.png,.jpg}
\usepackage{placeins}
\usepackage{caption}
\usepackage{subcaption}

\usepackage{booktabs}
\usepackage[table,xcdraw]{xcolor}
\usepackage{graphicx}
\usepackage{adjustbox}
\usepackage{multirow}

\usepackage{amssymb}
\usepackage{gensymb}
\usepackage[mathscr]{euscript}
\usepackage{bm}
\usepackage{mathtools}
\usepackage{nccmath}
\usepackage{amsmath}
\usepackage{braket}

\usepackage{algorithm}
\usepackage{algpseudocode}

\usepackage{lineno, hyperref}

\usepackage{lipsum}

\journal{International Journal of Applied Earth Observation and Geoinformation}

\begin{document}

\begin{frontmatter}

    \title{SuperpixelGraph: Semi-automatic generation of building footprint through semantic-sensitive superpixel and neural graph networks}

    \author{Haojia Yu}
    \author{Han Hu\corref{cor1}}
    \author{Bo Xu}
    \author{Qisen Shang}
    \author{Zhendong Wang}
    \author{Qing Zhu}
    \cortext[cor1]{Corresponding Author: han.hu@swjtu.edu.cn}

    \address{Faculty of Geosciences and Environmental Engineering, Southwest Jiaotong University, Chengdu, China}
    \begin{abstract}
    Most urban applications necessitate building footprints in the form of concise vector graphics with sharp boundaries rather than pixel-wise raster images. This need contrasts with the majority of existing methods, which typically generate over-smoothed footprint polygons. Editing these automatically produced polygons can be inefficient, if not more time-consuming than manual digitization. This paper introduces a semi-automatic approach for building footprint extraction through semantically-sensitive superpixels and neural graph networks. Drawing inspiration from object-based classification techniques, we first learn to generate superpixels that are not only boundary-preserving but also semantically-sensitive. The superpixels respond exclusively to building boundaries rather than other natural objects, while simultaneously producing semantic segmentation of the buildings. These intermediate superpixel representations can be naturally considered as nodes within a graph. Consequently, graph neural networks are employed to model the global interactions among all superpixels and enhance the representativeness of node features for building segmentation, which also enables efficient editing of segmentation results. Classical approaches are utilized to extract and regularize boundaries for the vectorized building footprints. Utilizing minimal clicks and straightforward strokes, we efficiently accomplish accurate segmentation outcomes, eliminating the necessity for editing polygon vertices. Our proposed approach demonstrates superior precision and efficacy, as validated by experimental assessments on various public benchmark datasets. A significant improvement of 8\% in AP50 was observed in vector graphics evaluation, surpassing established techniques. Additionally, we have devised an optimized and sophisticated pipeline for interactive editing, poised to further augment the overall quality of the results. The code for training the superpixel and graph networks will be made publicly available upon publication\footnote{\url{https://vrlab.org.cn/~hanhu/projects/spgraph/}}.
    \end{abstract}

\end{frontmatter}


\section{Introduction}
\label{s:intro}
Buildings are essential geographical features in urban environments. Accurate and up-to-date building footprints in the form of concise vector graphics are increasingly in demand for various applications, such as map services, urban planning, and reconstruction \citep{zhu2020interactive, haala2010update}. With the advancement of data acquisition techniques, high-resolution images with improved temporal and spatial resolution are now readily available. However, these images pose challenges for current building detection approaches, particularly in dealing with intricate roof designs and complex surrounding environments \citep{xu2021efficient, xiong2014graph}.

Two prevalent paradigms for generating building footprints currently exist: (1) pixelwise segmentation followed by regularization \citep{microsoft2018us}, and (2) end-to-end learnable approaches. The former utilizes semantic segmentation approaches \citep{long2015fully,ronneberger2015u} to create a binary probability map of buildings, followed by classical polygonal simplification and regularization methods for generating vector graphics. The latter directly predicts corner locations and connects them into closed polygons. Despite numerous impressive works on these paradigms, two issues remain unresolved:

(1) \textit{Excessive smoothing and inconsistent building boundaries}. Segmentation-based approaches generally produce oversmoothed building probability maps due to the inherent receptive field and multi-scale nature of convolutional neural networks (CNN) \citep{he2016deep}. The direct end-to-end approach \citep{zorzi2022polyworld,wei2023buildmapper} frequently creates irregular shapes that deviate from human preferences, which can be attributed to the learning procedure's generalization ability in predicting corner linkages.

(2) \textit{Automatically extracted building footprints can be inefficient for subsequent interactive quality control}. Despite commonly achieving 90\% precision metrics in segmentation, building footprints that have been over-smoothed or irregular shapes can be challenging to efficiently edit, thereby limiting their widespread use in real-world applications. Currently, editing operations typically involve direct manipulation of polygon vertices, such as moving, adding, and deleting, which can be extremely time-consuming.

\begin{figure}[H]
\centering
\subcaptionbox{Semantically-sensitive superpixels.}{\includegraphics[width=0.48\textwidth]{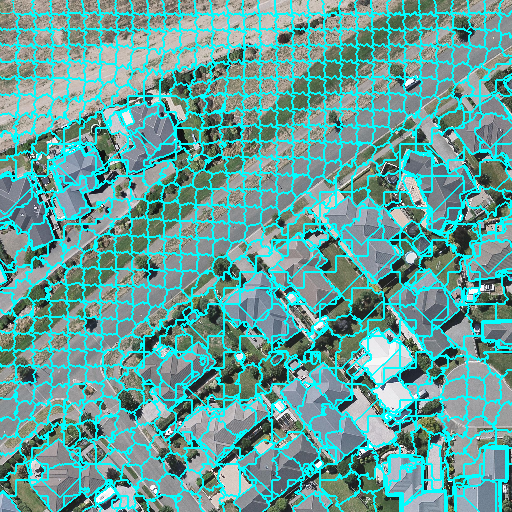}}
\subcaptionbox{Superpixel graph networks.}{\includegraphics[width=0.48\textwidth]{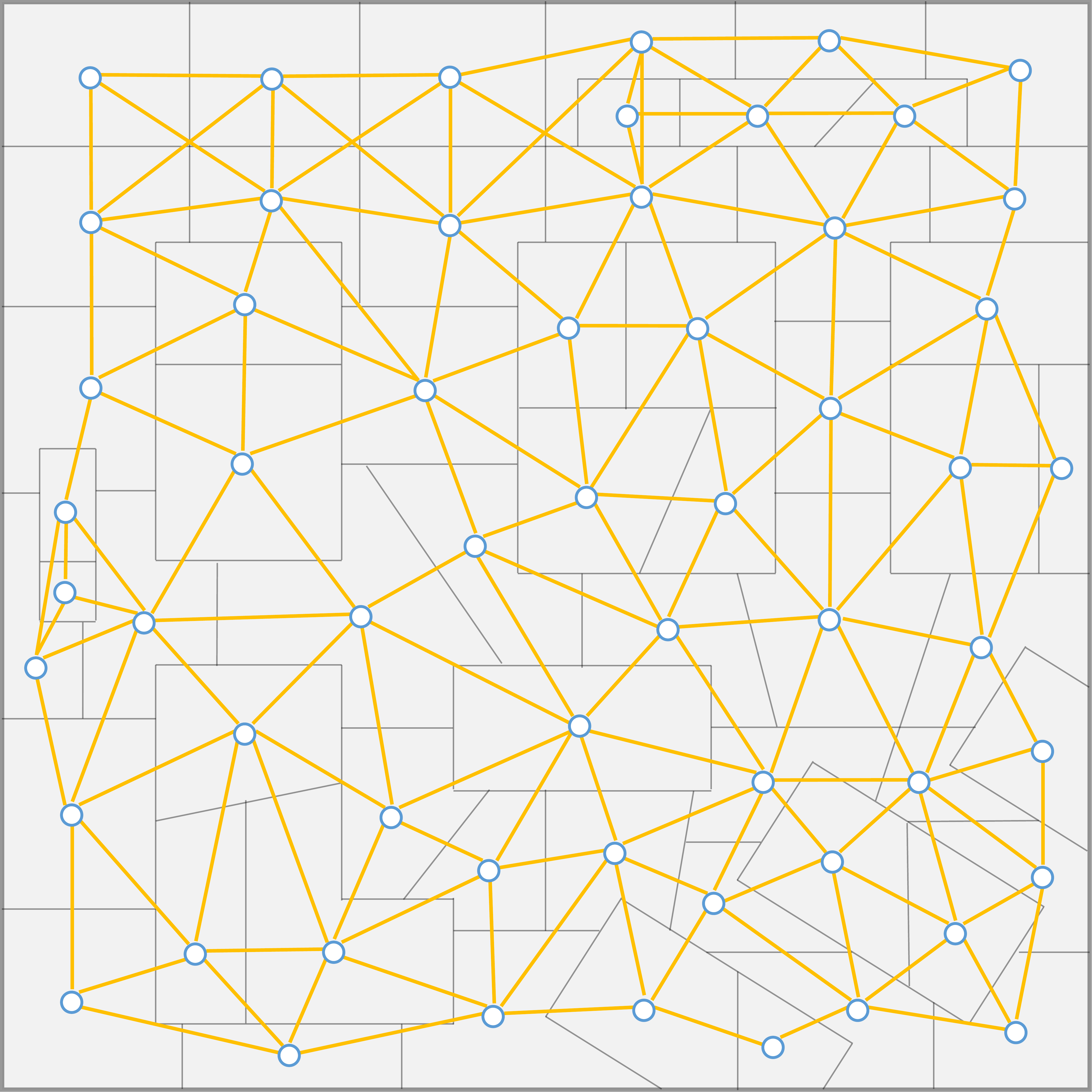}}
\subcaptionbox{Local addition (blue) and deletion (red) strokes with global effects. The affected nodes are highlighted in the corresponding colors.}{\includegraphics[width=0.96\textwidth]{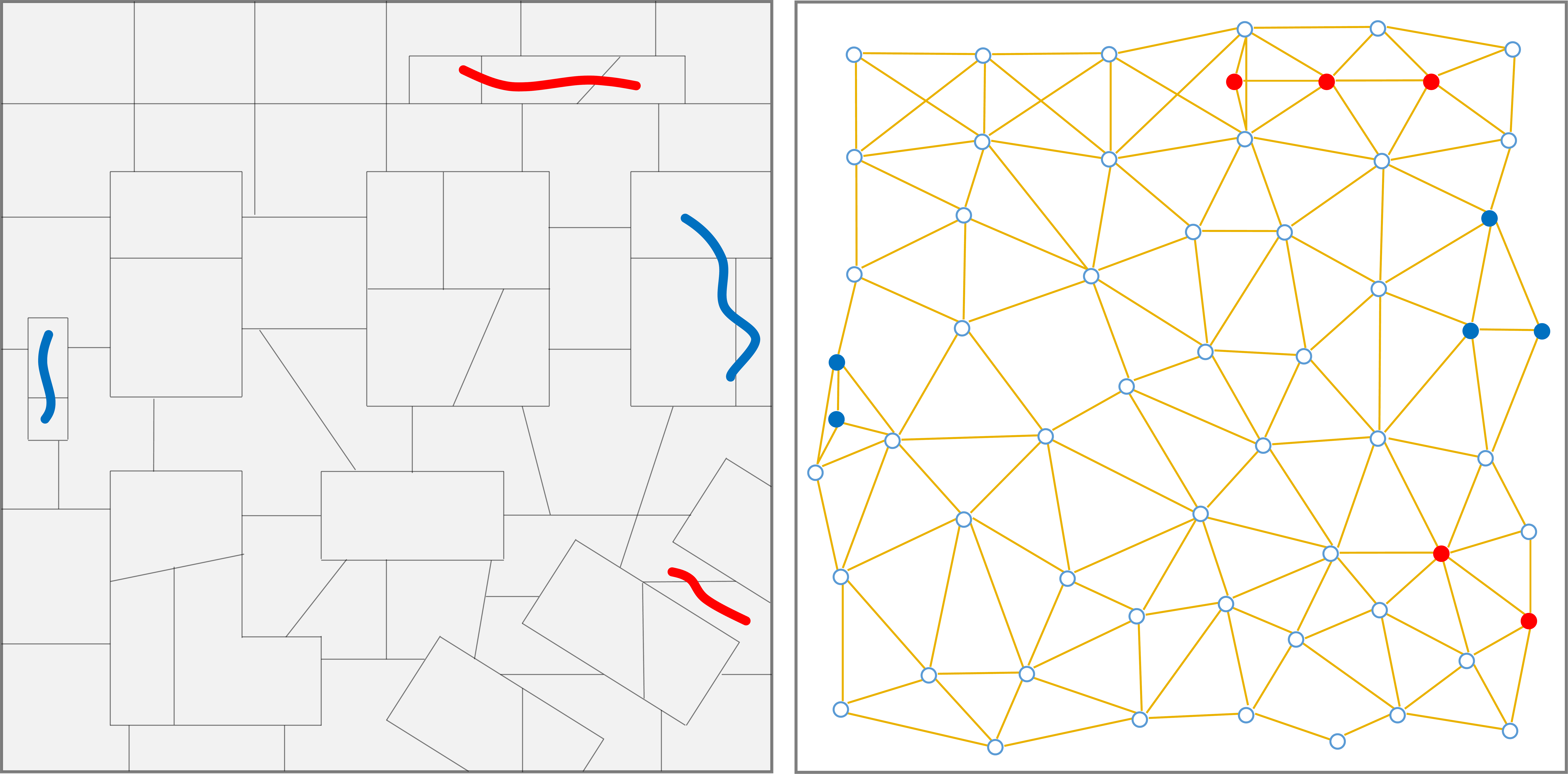}}
\caption{SuperpixelGraph enables improved boundary preservation, global optimization, and efficient editing.}
\label{fig:superpixel_graph}
\end{figure}

To address these issues, this paper proposes SuperpixelGraph for the semi-automatic generation of building polygons. We learn semantically-sensitive superpixels that react only to building boundaries, thus achieving improved boundary metrics (Fig. \ref{fig:superpixel_graph}a). Superpixels naturally partition images into globally connected graphs (denoted as SuperpixelGraph, Fig. \ref{fig:superpixel_graph}b), where node features and edge relationships are learned through graph neural networks. Owing to boundary preservation and global edge knowledge, local editing of SuperpixelGraph with addition and deletion strokes has global effects (Fig. \ref{fig:superpixel_graph}c). More specifically, we learn both superpixel clustering and semantic segmentation of buildings using the same encoder-decoder architecture, i.e., Fully Convolution Networks \citep{long2015fully}. The pixelwise feature maps are adaptively pooled into each superpixel, generating the original node feature of the SuperpixelGraph. Graph attention networks (GAT) \citep{velivckovic2017graph} are employed to embed node features and model pairwise potentials of superpixels. Segmentation is performed for each embedded feature of superpixels to generate rasterized building masks. Classical approaches \citep{suzuki1985topological,dyken2009simultaneous} are used to extract vector graphics of building footprints.

In summary, the contributions of the proposed method are twofold: (1) a multi-task network is designed to preserve building boundaries directly, effectively segmenting superpixels and building semantics simultaneously, and (2) the construction of a superpixel graph enhances the ability to distinguish between buildings and backgrounds while reducing the difficulty of human interaction. The remainder of the paper is organized as follows. Section \ref{s:relwork} discusses related work, while Section \ref{s:meth} presents the proposed methods in detail. Section \ref{s:expEva} comprises assessment and discussion. Finally, Section \ref{s:con} concludes the paper.

\section{Related work}
\label{s:relwork}

In the following sections, we focus on the most relevant topics related to this paper: (1) building footprint generation, (2) superpixels, and (3) graph neural networks.

\subsection{Building Footprint Generation}

\textbf{Raster building segmentation}. The advent of deep learning, particularly the proliferation of encoder-decoder architectures, has led to the adoption of semantic segmentation approaches for generating raster building masks. Multi-scale structures, such as UNet \citep{ronneberger2015u}, FCN \citep{long2015fully}, and FPN \citep{lin2017feature}, have enabled end-to-end training of binary building masks, significantly improving the practicality of such approaches. Buildings, as typical man-made structures, exhibit sharp edges. However, the hierarchical enlarged receptive field of CNNs, despite their impressive power in semantic modeling, inevitably smooths features over boundaries \citep{li2021building, liu2022building}. Many dedicated building segmentation studies focus on enhancing representativeness with respect to boundaries. Popular approaches include emphasizing features in shallow layers, auxiliary feature maps, and augmenting with low-level features \citep{zhu2020map, liao2021joint, zhou2022bomsc, jung2021boundary}. Nonetheless, the contradiction between an effective receptive field and precise boundary information remains unresolved.

\textbf{Footprint simplification and regularization}. Due to the concise data structure and support for spatial analyses, many applications require building representations in vector graphic formats. The conversion between raster and vector graphics has been extensively studied within the GIS and computer graphics communities, with a focus on polyline simplification and regularization. Polylines can be directly traced from binary masks using marching squares, forming the vectorized contour of objects \citep{suzuki1985topological}. By applying certain geometrical criteria such as length, distance, and angle, the number of vertices in the contour polylines can be reduced \citep{dyken2009simultaneous, gribov2019optimal}. To enforce shape regularities, strategies such as line fitting, discrete model selection, and least-squares optimization are commonly adopted \citep{dos2020regularization, xie2018hierarchical}. However, the quality of simplified and regularized results is inevitably affected by the initial raster mask. Nuisances in the raster mask, such as oversmoothing and zigzag issues, are generally difficult to remedy at this stage.

\textbf{Learned vectorized footprint}. Recently, a new paradigm has emerged for generating building footprints through directly learning vector generation. Various strategies have been attempted, including recurrent prediction of polygon vertices (RNN-based approaches) \citep{huang2021oec}, CurveGCN-like graph convolution networks along 1D curves \citep{ling2019fast,peng2020deep}, and learning offsets with initial object detection \citep{zorzi2022polyworld, wei2023buildmapper}. Moreover, extracting polygons with the assistance of auxiliary layers has demonstrated promising results, such as attractive field maps \citep{li2021building} and polyvector flow fields \citep{girard2021polygonal}. However, creating such layers still requires features in high-level semantic contexts, leading to a gap between pixel-wise features and high-level semantic features. Additionally, the generalization ability is also a concern when learning to connect the corners of buildings.

\subsection{Superpixels}

\textbf{Classical Clustering-based Superpixels}. Superpixels are a partition of images into over-segments that should not cross the boundaries of objects of interest \citep{stutz2018superpixels,achanta2012slic}. By confining superpixels within the same objects, we can classify the region belonging to a single superpixel as a whole, thus improving the robustness and sharpness of semantic classification. This technique is also known as Object-based Image Analysis \citep{blaschke2010object}. Most classical methods for generating superpixels rely on clustering low-level features, such as intensity or gradient values of images. Clustering approaches include watershed segmentation \citep{benesova2014fast}, mean shift \citep{comaniciu2002mean,vedaldi2008quick}, geometric flows \citep{levinshtein2009turbopixels}, K-means \citep{achanta2012slic,li2015superpixel}, graph cuts \citep{veksler2010superpixels}, and others. However, it is important to note that superpixels should only be confined to the boundaries of \textit{interested objects}, rather than any object. For example, if we are only interested in buildings, allowing superpixels to cross roads may also be beneficial to the final outputs (Fig. \ref{fig:superpixel_graph}a).

\textbf{Learned approach for superpixels}. The hard assignment of each pixel to the corresponding superpixel is apparently not differentiable, making it nearly impossible to learn to generate superpixels. This challenge was addressed by the seminal work of the Superpixel Sampling Network (SSN) \citep{jampani2018superpixel}, which proposed an elegant design for the soft connection matrix. SSN introduced a differentiable SLIC-like K-means iteration step as a workaround. Following a similar pipeline, another study proposed an FCN architecture to avoid the iteration of the differentiable SLIC by directly learning the soft connection matrix \citep{yang2020superpixel,zhu2021learning}. However, these approaches consider all natural boundaries, posing difficulties in improving the recall rate of the interested objects \citep{ng2023fuzzy}. In this paper, we adopt an additional semantic branch to make the superpixel generation semantically sensitive.

\subsection{Graph Neaural Networks}

Graph Neural Networks (GNNs) have emerged as a powerful framework for learning and representation of structured data in the form of graphs. Over the past decade, GNNs have experienced significant growth and have been applied to a wide range of problems. The foundational work of \cite{scarselli2008graph} introduced the concept of Graph Neural Networks, which paved the way for various GNN architectures, such as Graph Convolutional Networks (GCNs) \citep{kipf2016semi}, Graph Attention Networks (GATs) \citep{velivckovic2017graph}, and Graph Isomorphism Networks (GINs) \citep{xu2018powerful}. These architectures have been designed to leverage both local and global information in graphs by iteratively aggregating and transforming neighbor node features. Despite the embedding of features attached to the nodes of the graphs, features can also be learned for the edge connections of the graph, representing the relationships between nodes.

Upon segmenting images into superpixels, it becomes intuitive to model the image as a graph. The features corresponding to each superpixel can be derived through global pooling operations within the superpixel. The primary distinction between the proposed methodology and prior works in image or point cloud classification lies in our approach to generate semantically-relevant superpixels. Additionally, our method places emphasis on both edges within the graph and the relationships they represent, as opposed to merely concentrating on node feature embeddings.

\section{Methodology}
\label{s:meth}
\subsection{Overview and problem setup}
\label{ss:overview}

In this manuscript, we present a sophisticated and pragmatic pipeline tailored for the generation of vectorized building polygons. While the pursuit of a fully automated pipeline remains the paramount goal, the significance of developing efficacious interactions should not be understated. Our findings reveal that manual digitization or modification of vertices, particularly those necessitating strict conformity to underlying orthophotos, is considerably laborious. Human operators exhibit exceptional skill in executing perceptually accurate yet geometrically imprecise tasks. Consequently, to achieve streamlined interactions, it is imperative to incorporate automatically generated auxiliary information characterized by the following attributes: (1) geometric segments exhibiting a high degree of alignment with the intended boundaries, and (2) information encapsulating global context.

Nonetheless, the aforementioned criteria pose an intrinsic contradiction. The former necessitates localized information, whereas the latter demands global context. This insight led us to partition the pipeline responsible for generating auxiliary information into two discrete phases. In the initial phase, the pipeline is designed to produce superpixels that adhere to building boundaries. The learning process harnesses SSN-like methodologies \citep{jampani2018superpixel}, fundamentally capturing pixel pair-wise similarities and inherently encoding local information. Subsequently, in the second phase, superpixel segmentation is treated as a graph, with global context being modeled through the utilization of graph neural networks. Node features are systematically and hierarchically amalgamated to encapsulate global context, while the interrelationships between these features are appraised correspondingly.

\begin{figure}[H]
  \centering
    \includegraphics[width=\textwidth]{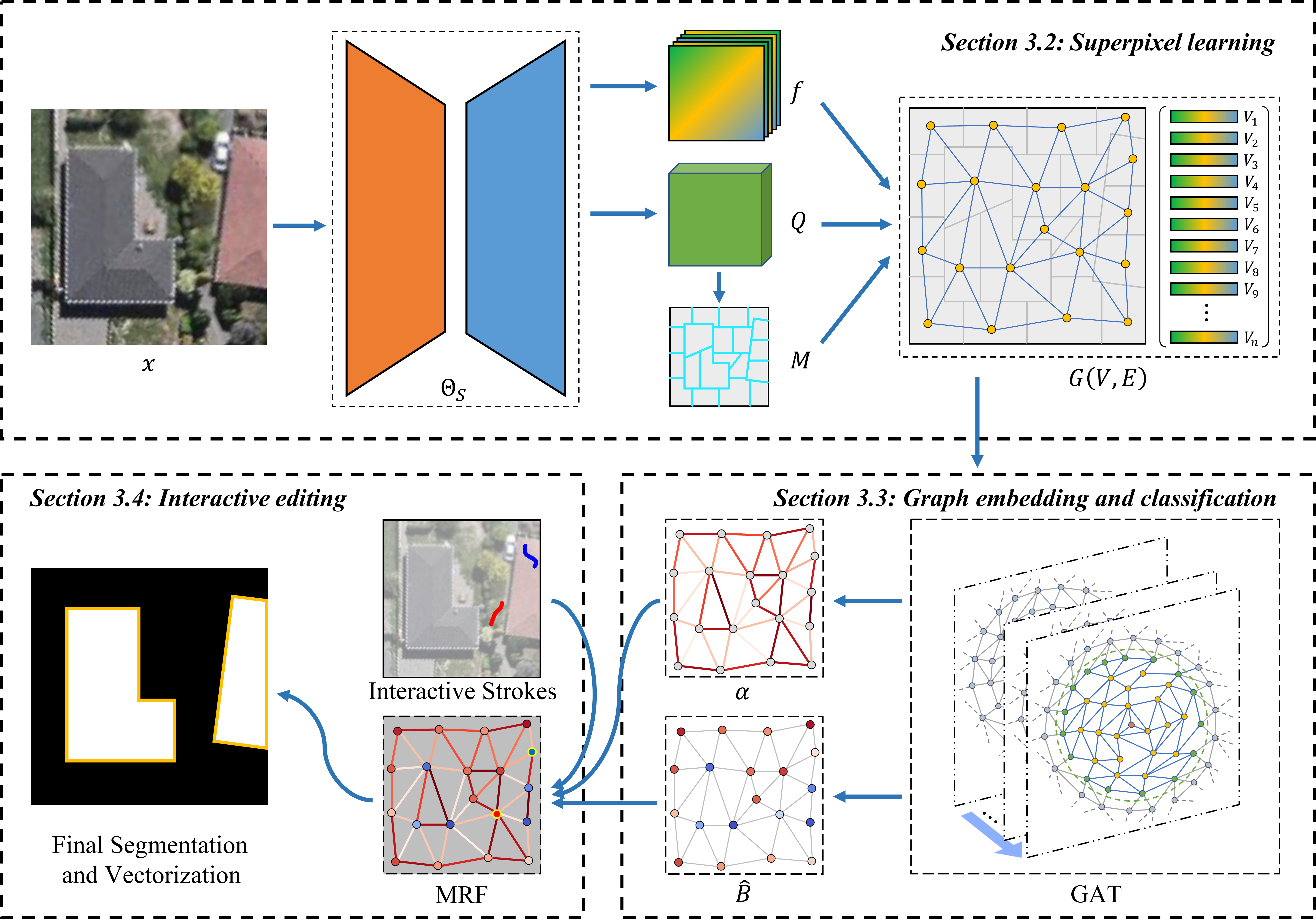}
	\caption{The overall workflow of the proposed approach.}
     \label{fig:overall}
\end{figure}

In a more formal context, the inputs comprise the color image $x\in\mathbb{R}^{S\times3}$ (with image size $S=W\times H$) and the corresponding binary building mask $b\in\mathbb{Z}^{S}$. Each pixel $b_p\in\left \{0, 1 \right \}$ signifies either the building or background. To render the superpixel learnable, the building mask $b$ is additionally transformed into the one-hot feature $h\in\mathbb{R}^{S\times 2}$. Our proposed approach, denoted as SuperpixelGraph, incorporates the following key steps to semi-automatically generate vectorized buildings (refer to Fig. \ref{fig:overall}):

\textbf{Semantically-Sensitive Superpixel Network} $\Theta_S(x)\rightarrow (Q,M,f)$. The initial stage yields auxiliary over-segmentation $M\in\mathbb{Z}^S$ of the image $x$ into $N$ superpixels\footnote{Henceforth, we generally denote pixel-wise features with lowercase letters and superpixel-wise features with uppercase letters.}, implying that each pixel of the segmentation $M_p\in \left \{0,1,2,\cdots,N-1 \right \}$. A conventional encoder-decoder network is employed to facilitate learning the superpixel in an end-to-end fashion, wherein a differentiable pixel-superpixel association matrix $Q\in\mathbb{R}^{S\times 9}$ \citep{jampani2018superpixel} and a $C$-channels feature map $f\in\mathbb{R}^{S\times C}$ are concurrently generated. Each pixel in $Q$ represents the probability that pixel $p$ belongs to one of the adjacent $9$ superpixels. The $N$ superpixels, as opposed to the individual $S$ pixels, are utilized for classification. Consequently, the superpixels ought to be sensitive to building semantics. Further details are elaborated upon in Subsection \ref{subs:superpixel_network}.

\textbf{SuperpixelGraph} $(Q, M, f) \rightarrow G(V, E)$. The construction of SuperpixelGraph $G(V,E)$, derived from the superpixels $M$ and feature maps $f$, is rather straightforward. The vertices $V$ are endowed with the corresponding node feature $V\in\mathbb{R}^{N\times C}$. The node feature is produced by implementing a weighted average utilizing $Q$ and $f$. The edge indices $E\in\mathbb{Z}^{|E|\times 2}$ are acquired by tracing adjacent pixels in segmentation $M$ and gathering boundary indices.

\textbf{SuperpixelGraph Embedding and Classification} $\Theta_G(G)\rightarrow (V^L, \alpha)$. A graph neural network $\Theta_G$ incrementally aggregates and projects the node feature into the final layer $V^L\in\mathbb{R}^{N\times C}$ ($L$ represents the number of layers). Subsequently, the features $V^L$ are classified into superpixel-wise probabilities for the building $\hat{B}\in\mathbb{R}^{N}$. Moreover, $\Theta_G$ concurrently generates similarities $\alpha\in\mathbb{R}^{|E|}$ for adjacent nodes, as described in Subsection \ref{subs:gat}.

\textbf{Globally Optimized Interactive Editing}. The graph $G$ can also be construed as a Markov Random Field (MRF), wherein probability $\hat{B}$ exemplifies the data term, and node similarities $\alpha$ for the smooth term. Editing operations, encompassing addition and deletion, can be represented by strokes on the segmentation $M$. Probabilities $\hat{B}$ with indices in $M$ that overlap with strokes are enforced to 1 or 0 for addition and deletion, respectively. A global label optimum is attained through the graph cut.

\textbf{Building Footprint Vectorization}. After editing, the probabilities of the superpixel nodes are rendered to a binary mask. Classical raster-to-vector conversion methods \citep{suzuki1985topological} with polygon simplification and regularization \citep{dyken2009simultaneous,xie2018hierarchical} are adopted to generate the vectorized building footprint.

\subsection{Semantically-Sensitive Superpixel Network}
\label{subs:superpixel_network}

This paper adopts the SSN strategy \citep{jampani2018superpixel} to learn features for generating superpixels. The core of SSN lies in establishing the pixel-superpixel association matrix $Q$, where each pixel $Q_p\in\mathbb{R}^{9}$ denotes the probability that pixel $p$ belongs to one of the adjacent $9$ potential superpixels. Utilizing $Q$, we can directly aggregate the superpixel feature from the pixel-wise feature map and also disperse the superpixel feature back to the pixel-wise feature through weighted interpolation. To make this process learnable, appearance reconstruction and size regularities of the superpixel are used as loss functions. This strategy enforces intra-consistency of the image's appearance information, which is sensitive to various appearance changes along boundaries. To further enhance boundary preservation efficacy for buildings, this paper introduces an additional semantic branch. The goal is to make the superpixel solely responsive to building boundaries, thereby mitigating potential influences from other objects.

\begin{figure}[H]
  \centering
    \includegraphics[width=\textwidth]{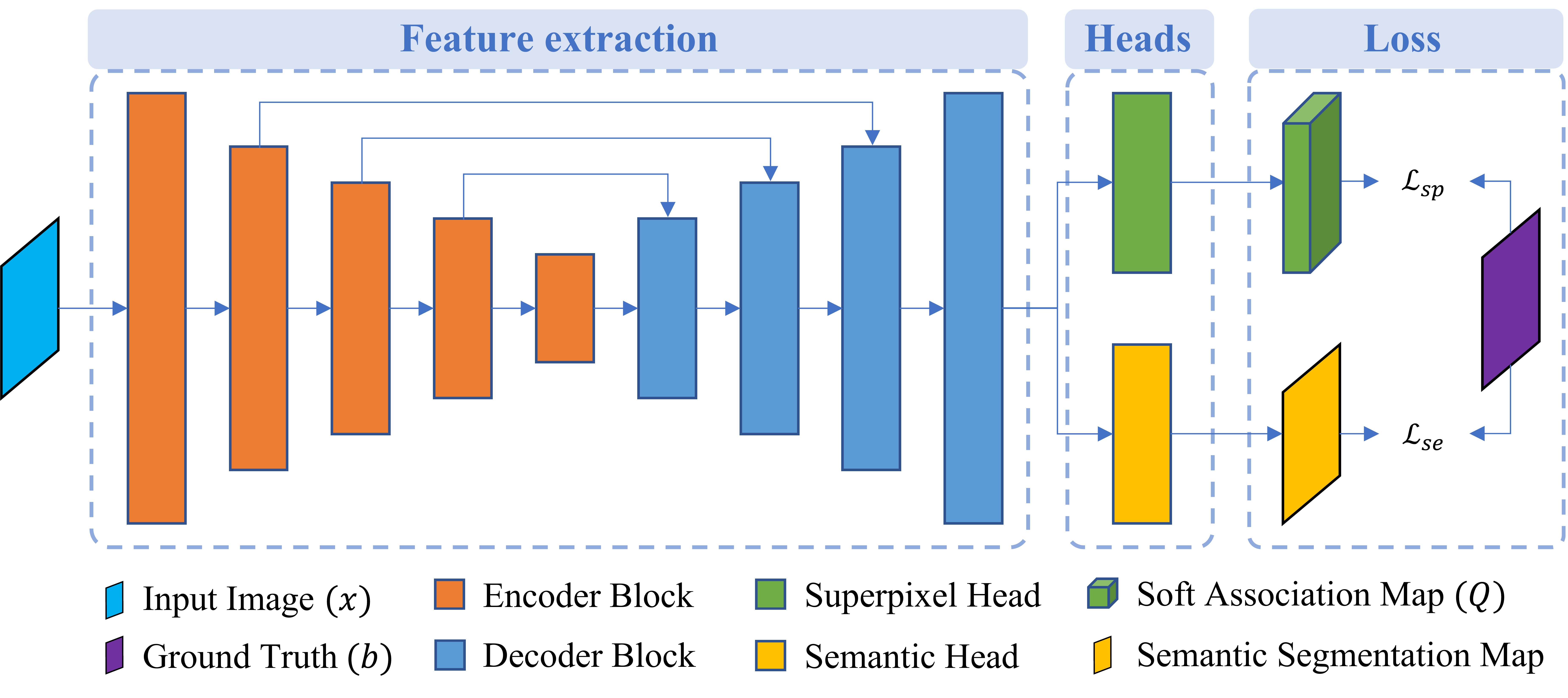}
	\caption{The architecture of the semantically-sensitive superpixel generation network.}
   \label{fig:superNetwork}
\end{figure}

The architecture of the network (Fig. \ref{fig:superNetwork}) for the generation of semantically-sensitive superpixels is quite straightforward, which consists of a feature extractor, two heads for superpixel and semantic segmentation and the loss functions. The core strategy for the network is inspired by SSN. For completeness, we also briefly introduce the necessary components of SSN for better understanding of this paper.

\textbf{Feature Extractor}. We adopt a standard encoder-decoder architecture to generate a pixel-wise feature map $f$ from the input image $x$, e.g., $\Theta_F(x)=f$. In the encoder part, we use a structure similar to nested U$^2$-Net \citep{qin2020u2}, with each encoder block having a Residual U-block. U$^2$-Net is experimentally found to better capture subtle information along object boundaries for salient detection, particularly for thin objects, making it suitable for our purpose. In this way, each layer can combine information extracted from multiple scales, allowing the encoder to capture more context.

\textbf{Superpixel and Segmentation Heads}. The feature map $f$ branches into two distinct paths: the pixel-superpixel association matrix $Q$ in the superpixel heads, and the esimated building segmentation map $\hat{b}\in\mathbb{R}^{S \times 2}$ in the semantic segmentation heads. The weights are essentially convolutions with a $1\times 1$ kernel.

\begin{figure}[H]
    \centering
    \includegraphics[width=0.5\linewidth]{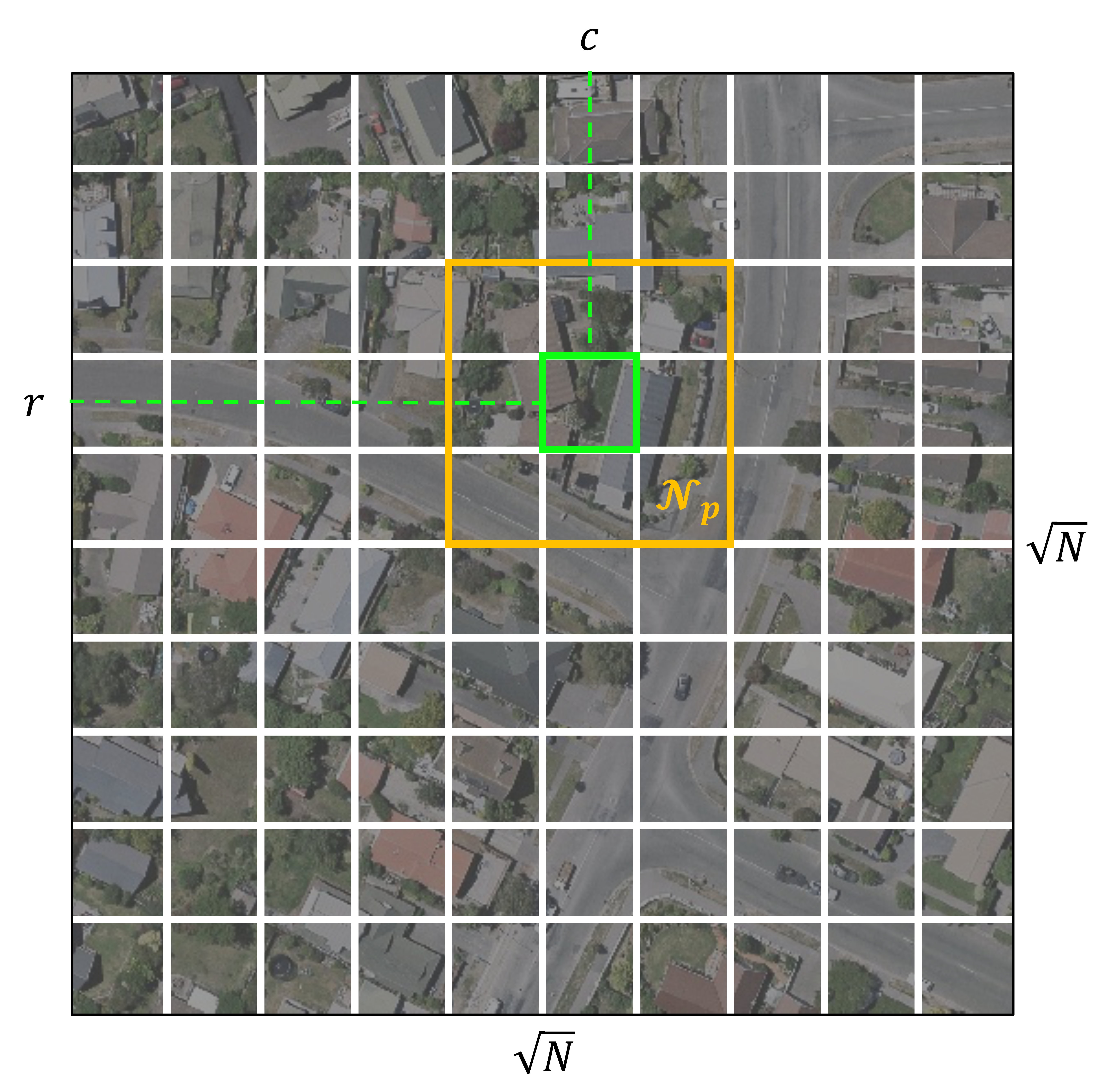}
    \caption{Fixed lattice grid and definition of the adjacent neighborhood.}
    \label{fig:superpixelNeighbor}
\end{figure}

As illustrated in Fig. \ref{fig:superpixelNeighbor}, the image is initially partitioned into $N$ square lattice grids, with $\sqrt{N}$ cells allocated to each dimension. Each pixel in the association matrix $Q_p$ aims to estimate the probability of belonging to the adjacent grids $\mathscr{N}_p$ corresponding to pixel, which is determined as follows:

\begin{equation}
\mathscr{N}_p=\{ (r+\Delta r) \times \sqrt{N} + (c+\Delta c) | \Delta r, \Delta c\in[-1,0,1]\}
\end{equation}
where $(r, c)$ represents the row and column of the grid that pixel $p$ is located in. For cells situated along the image border, the cells are padded with reflected values.

\textbf{Superpixel Aggregation and Dispersion}.
Given the pixel-superpixel association matrix $Q$, we can softly aggregate pixel-wise features into superpixel-wise features. In this paper, we use the pixel-wise one-hot feature $h$ for the binary building mask and convert it to the superpixel-wise representation $H \in \mathbb{R}^{N \times 2}$. Since only the adjacent $9$ superpixels are relevant, we sparsely estimate aggregation using adjacent grids. The row and column of the $n$-th superpixel are determined by $n = r \times \sqrt{N} + c$. In addition to the aggregation operator, we can also reconstruct the pixel-wise feature $\hat{h}$ from the aggregated superpixel-wise feature $H$ using weighted interpolation. For the $n$-th superpixel, the aggregated feature $H_n$ and the dispersed feature $\hat{h}_p$ are computed as follows:
\begin{equation}
\begin{split}
H_n &= \frac{1}{Z_n} \sum_{p \in \mathscr{N}_p} h_p \cdot Q_p(n) \\
\hat{h}_p &= \sum_{n \in \mathscr{N}_p} H_n \cdot Q_p(n)
\end{split}
\label{eq:agg_sem}
\end{equation}
where $Q_p(n)$ indicates the probability of the $n$-th superpixel in $Q_p$, and $h_p \in \mathbb{R}^2$ is the one-hot feature for pixel $p$ in the building map. The aggregation only considers pixels inside the adjacent lattice grids $\mathscr{N}p$. $Z_n$ is a normalizer for the corresponding superpixel:
\begin{equation}
Z_n = \sum_{p \in \mathscr{N}_p} Q_p(n)
\end{equation}

In addition to the aggregation and dispersion for the building masks $h$, the same strategy is also applicable for the pixel location $p$ and the center of the superpixel $P$,
\begin{equation}
\begin{split}
P_n&=\frac{1}{Z_n}\sum_{p\in\mathscr{N}_p} p\cdot Q_p(n) \\
\hat{p} &=\sum_{n\in\mathscr{N}_p} P_n \cdot Q_p(n)
\end{split}
\label{eq:agg_pos}
\end{equation}
where $\hat{p}$ is the reconstructed location from superpixels. The location is crucial for maintaining the regular shape of the superpixel.

\textbf{Loss Functions}. The loss function comprises two components: the superpixel loss $\mathscr{L}_{sp}$, which enforces intra-consistency and shape regularity within each superpixel, and the semantic loss $\mathscr{L}_{se}$, which facilitates learning features that distinguish buildings from the background. For the superpixel loss, supervision is provided by the original one-hot vector for buildings and the reconstructed one derived from the pixel-superpixel association matrix $Q$, taking into account shape regularities:
\begin{equation}
\mathscr{L}_{sp} = CE(h, \hat{h}) + \lambda \sum_{p} || p - \hat{p} ||
\end{equation}
where $CE(\cdot, \cdot)$ denotes the cross-entropy and $\lambda$ serves as a weight to balance the regularization term. For the semantic loss, the conventional approach employed in semantic segmentation is utilized:
\begin{equation}
\mathscr{L}_{se} = CE(b, \hat{b})
\end{equation}
The final loss is a direct combination of these two components.

\textbf{Superpixel Clustering and Feature Pooling}. To generate the superpixel clustering map $M$, we compute the hard pixel-superpixel association directly by taking the maximum in the soft association matrix $Q$.
\begin{equation}
M_p = \arg\max_{n \in \mathscr{N}_p} Q_p(n)
\end{equation}
The corresponding feature for the $n$-th superpixel, $V_n \in \mathbb{R}^{C}$, is aggregated using the feature map $f$, in a manner similar to Equations (\ref{eq:agg_sem}) and (\ref{eq:agg_pos}).
\begin{equation}
V_n = \frac{1}{Z_n} \sum_{p \in \mathscr{N}_p} f_p \cdot Q_p(n)
\label{eq:agg_v}
\end{equation}

\subsection{SuperpixelGraph Embedding with Graph Attention Networks}
\label{subs:gat}

The aggregated superpixel feature $V \in \mathbb{R}^{N \times C}$ is intrinsically local. $\mathscr{L}_{sp}$ takes into account only the intra-consistency within each superpixel. Although the encoder-decoder architecture progressively expands the receptive field of the CNN, and $\mathscr{L}_{se}$ leverages semantic information, the features must capture a broader context to achieve robustness in semantic segmentation. Moreover, as only intra-consistency is considered, the relationships between adjacent superpixels remain ambiguous. Consequently, the embedding step for SuperpixelGraph serves two primary objectives: (1) propagating and aggregating contextual information throughout the graph, and (2) explicitly estimating the relationships between superpixels, thereby enriching the information on the edges of the graph.

\textbf{Graph Embedding Network}.
Extending the convolution operator to irregular domains typically involves a neighborhood aggregation or message passing scheme \citep{fey2019fast}. In this paper, a vertex represents a superpixel with its node feature denoted by $V_n \in \mathbb{R}^{C}$. The message passing formulation is employed to propagate information progressively to the target node through multiple layers $l$ along the neighborhood. Specifically, a single layer of the message passing graph neural networks can be described as follows:

\begin{equation}
V_i^l = \Theta_{1}^l \left( V_i^{l-1}, \underset{(i,j)\in E}{\oplus} \Theta_2^l \left( V_i^{l-1}, V_j^{l-1} \right) \right)
\end{equation}

In this context, the superscript $l$ signifies the number of layers, while the subscripts $(i,j)$ denote the indices of nodes. The symbol $\oplus$ represents an order-invariant function, such as sum or max. $\Theta$ refers to learnable functions, typically Multi-Layer Perceptrons (MLPs) with activation and normalization functions. However, the concept of message passing is general, and other modular schemes are also possible. In this paper, we leverage Graph Attention Networks (GAT) \citep{velivckovic2017graph} for the message passing layer.

\begin{figure}[H]
\centering
\includegraphics[width= \textwidth]{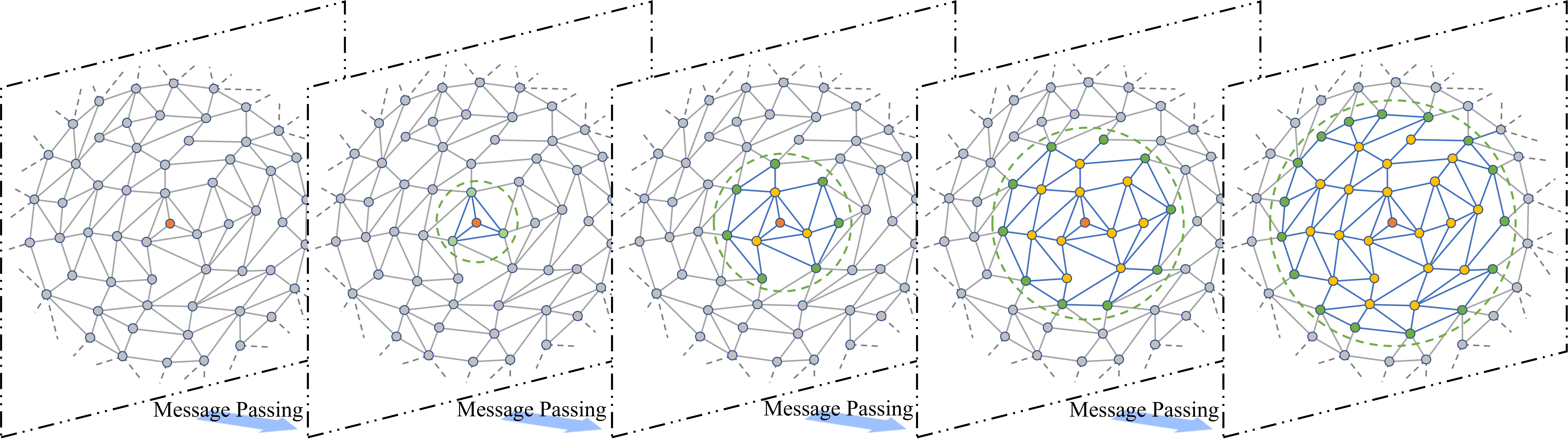}
\caption{SuperpixelGraph employs a multitude of Message Passing layers to assimilate contextual information effectively. The targeted node is signified by a red dot, while the green dots represent the corresponding nodes incorporated in the current layer. The yellow dots, on the other hand, denote the aggregate nodes contemplated within the Message Passing networks.}
\label{fig:graphMessage}
\end{figure}

As depicted in Fig. \ref{fig:graphMessage}, this paper employs four message passing layers. As detailed in Section \ref{ss:overview}, the construction of the SuperpixelGraph considers only direct neighbors for edge links. Each layer propagates and aggregates information in a ring structure to the target node. Consequently, information along the edges with the 4-th degree becomes visible to the target node. Incorporating more context information can be achieved by stacking additional layers; however, this may also introduce irrelevant information, thereby compromising performance. We find that four layers strike an optimal balance between contextual information and precision.

\textbf{Graph Attention Networks}.
GAT represents an instantiation of the message passing layer and is formulated as follows:
\begin{equation}
V_i^l=\alpha_{ii}\Theta(V_i^{l-1}) + \sum_{(i,j)\in E}\alpha_{ij}\Theta(V_j^{l-1})
\end{equation}
where $\Theta$ is a shared MLP for each layer. $\alpha_{ij}$ is the attentional score between nodes $(i,j)$. In contrast to classical attention maps that model dense connections between all nodes, $\alpha_{ij}$ only considers the directly linked nodes $(i,j)\in E$ in the graph, making it sparse. The attention coefficients $\alpha_{ij}$ are computed using a modified dot attention mechanism:
\begin{equation}
\alpha_{ij}=\frac
{\exp \left( \sigma( \mathbf{a} \cdot [ \Theta(V_i) || \Theta(V_j)] ) \right)}
{ \sum_{(i,k)\in E} \exp \left( \sigma( \mathbf{a} \cdot [ \Theta(V_i) || \Theta(V_k)] ) \right) }
\label{eq:attScore}
\end{equation}
where $\mathbf{a} \in \mathbb{R}^{2C}$ is the learned vector; the operator $[\cdot||\cdot]$ concatenates two node features; and $\sigma$ is the activation function, such as LeakyReLU \citep{xu2015empirical}, which is used in this paper.

\begin{figure}[H]
\centering
\begin{subfigure}[b]{\linewidth}
    \includegraphics[width= \textwidth]{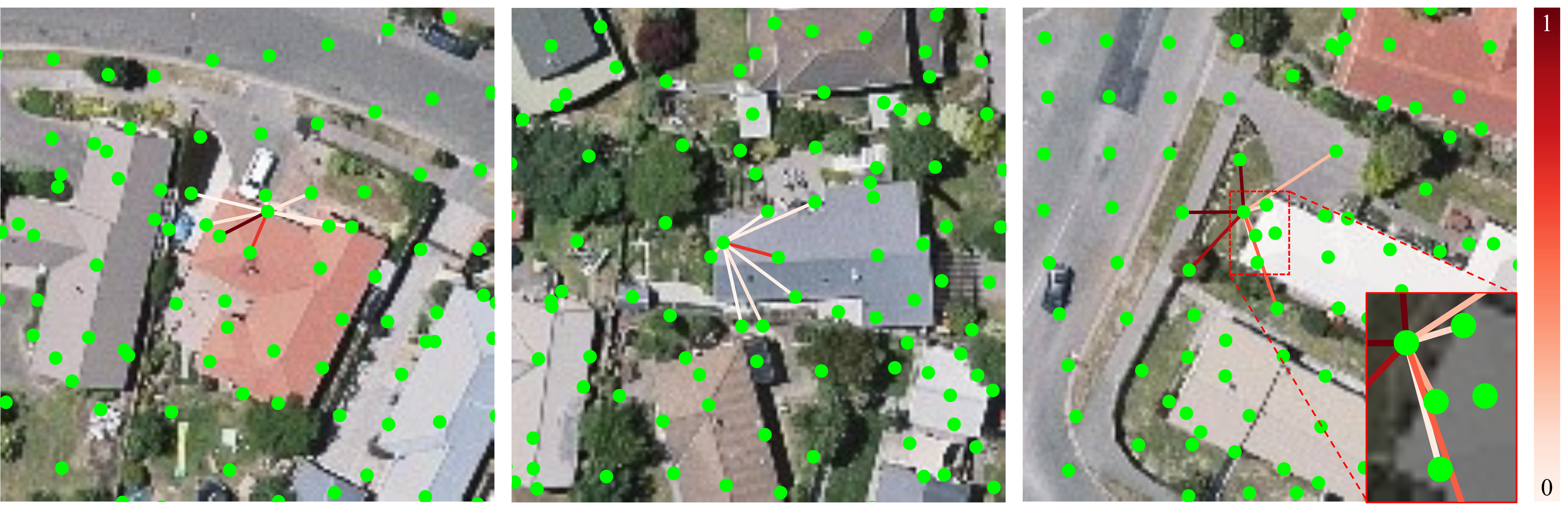}
\end{subfigure}
\caption{Visualization of the attention score for a specific node. Attentional aggregation constructs a dynamic graph between adjacent superpixels. Weights $\alpha_{ij}$ are shown as segments, with normalized values represented by colors.}
\label{fig:attWeights}
\end{figure}

As shown in Equation (\ref{eq:attScore}), $\alpha_{ij}$ essentially represents a learnable correlation between adjacent node features $(V_i, V_j)$, which is normalized using the $softmax$ function among all nodes connected to $V_i$. As depicted in Fig. \ref{fig:attWeights}, superpixels belonging to the same objects exhibit a markedly higher correlation compared to those of inter-object connections. This observation inspired us to utilize the attention scores for subsequent interactive editing.

\textbf{Loss functions}.
The final layer of the above embedding yields the logits for the classification of the superpixel, denoted as $V^4$. The corresponding label mask $b$ should also be aggregated to the dynamically created graph. The superpixel-level label $B\in\mathbb{Z}^{N}$ is generated using scattered reduction from the dynamically created superpixel segmentation $M$ and the mask $b$. Cross entropy is employed as the loss function $\mathscr{L}_G$ to learn the weights for the graph embedding and attentional aggregation.

\begin{equation}
\mathscr{L}_G=CE(B, V^4)
\end{equation}

\subsection{Stroke editing with global optimization}
In the inference step, the outputs of the SuperpixelGraph consist of the superpixel segmentation map $M$, the estimated superpixel-level probability of building segmentation $\hat{B}$, and the correlation scores $\alpha$. In practical applications, incorrect classification of buildings is inevitable. Therefore, we design an efficient global optimization step that takes into account user interaction.

\begin{figure}[H]
\centering
\includegraphics[width=\linewidth]{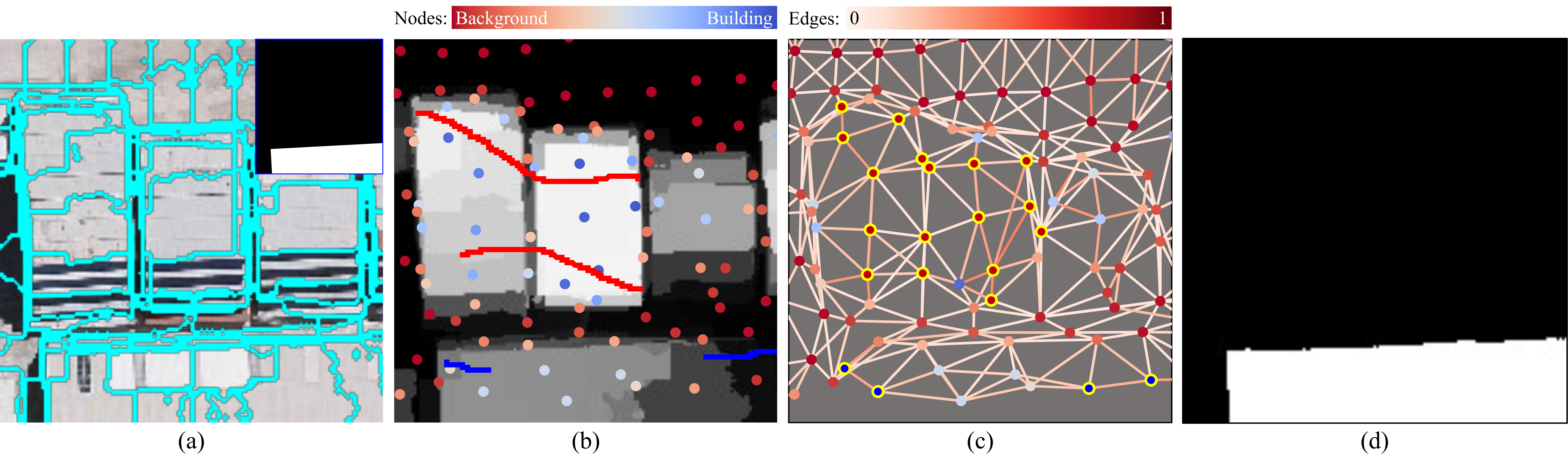}
\caption{Interactive stroke editing with global optimization. (a) The image overlaid with superpixels, with the edges of the superpixels highlighted in cyan. The ground-truth mask is aligned with the upper-right corner of the image. (b) The automatically generated pixel-level building class probability map and the corresponding superpixel classes, where each circle represents a superpixel, with the closer it is to blue indicating a higher probability of being a building, and vice versa for non-building. The red and blue strokes represent labeling the corresponding regions as non-building and building, respectively. (c) The graph network optimized globally using the Markov Random Field (MRF) \citep{boykov2004experimental}. The edited nodes are highlighted by yellow outlines, with red and blue representing modification of the building class probability of the node to 0 and 1, respectively. (d) The optimized building segmentation mask obtained after the optimization process.}
\label{fig:mrfEditing}
\end{figure}

As depicted in Fig. \ref{fig:mrfEditing}, an intermediate building mask is created using hard thresholding from the segmentation score $\hat{B}$. The mask is overlaid on the image, allowing the operator to add or delete building regions using stroke editing. The probabilities in $\hat{B}$ for the corresponding superpixels that are affected by the editing strokes are reset to 0 and 1 for deletion and addition, respectively. Then, a Markov Random Field using the weighted Potts model is constructed, and the final label $L\in\mathbb{Z}^N$ for each superpixel is solved using graph cuts:

\begin{equation}
E(L)=\sum_{i<N}{1-\hat{B}_i(L_i)} + \varphi \sum_{(i,j)\in E}{\alpha_{ij}I(L_i, L_j)}
\end{equation}
where $I(\cdot,\cdot)$ is an indicator function that evaluates to 0 if the two labels are the same and 1 if not. $\varphi=10$ is the weight that balances the data term and smooth term.

After the graph cut-based optimization concludes, a precise building mask can be generated. The next task is to track the building outline and vectorize it based on parallel and vertical constraints. In this work, we adopt existing solutions \citep{dos2020regularization,xie2018hierarchical} to trace the boundary and regularize the vector graphics.

\section{Experiments and evaluation}
\label{s:expEva}

In this section, we conduct a comprehensive experimental evaluation of the proposed techniques. A brief overview of the data sets is initially presented in Section \ref{s:experimentsData}, followed by an assessment of the superpixel segmentation in Section \ref{s:superEva}. Subsequently, an in-depth analysis of the precision and efficiency of the building extraction, focusing on local details, is provided in Section \ref{s:preAndLoc}.

\subsection{Datasets}
\label{s:experimentsData}
Three datasets are employed in our experiments.
The first dataset, the WHU aerial dataset \citep{ji2018fully}, is sourced from New Zealand, encompassing 187,000 buildings within a 450 $km^2$ area. The dataset comprises 8,188 RGB image tiles in total, each containing 512$\times$512 pixels with a 0.3m spatial resolution. It is divided into three parts: training set, validation set, and testing set, which consist of 4,736, 1,036, and 2,416 tiles, respectively.
The second dataset, the INRIA aerial dataset \citep{maggiori2017can}, is gathered from 10 cities worldwide, covering an area of 810 $km^2$. Each city features 36 images with dimensions of 5000$\times$5000 and a spatial resolution of 0.3m. We employed only the training set of the dataset for quantitative experiments, cropping the images into 512$\times$512 with 12 pixels of overlap. The tiles for training were randomly split by an 8:2 ratio for the training set and validation set.
The third dataset, the Vegas satellite dataset, is a subset of the Deep Globe Building Extraction Challenge dataset \citep{demir2018deepglobe}, captured by the WorldView-3 satellites. This dataset consists of 3,851 images and 110,000 buildings. We selected Pan-sharpened RGB images with dimensions of 650$\times$650 and a resolution of 0.3m spatial. The ground truth binary labels were generated using the summary file containing the spatial coordinates of all annotated building footprint polygons. All images were randomly divided by a 6:1.5:2.5 ratio for the training, validation, and testing sets.
A comparison of the datasets is provided in Table \ref{tab:expDatasets}.

\begin{table}[h]
    \centering
    \caption{Summary of the three datasets used for evaluations.}
	\begin{tabular}{lcccc}
	\hline
	\multirow{2}{*}{Dataset} & \multirow{2}{*}{Type} & \multirow{2}{*}{Area(km$^2$)} & \multirow{2}{*}{Size} & Samples                 \\
							 &                       &                               &                       & (train/validation/test) \\ \hline
	WHU                      & Aerial                & 450                           & 512$\times$512        & 4736/1036/2416          \\
	INRIA                    & Aerial                & 810                           & 512$\times$512        & 12400/3100/3000         \\
	Vegas                    & Satellite             & 216                           & 650$\times$650        & 2310/577/964            \\ \hline
	\end{tabular}
    \label{tab:expDatasets}
\end{table}

\subsection{Evaluation of superpixel segmentation}
\label{s:superEva}
The performance of the proposed superpixel segmentation algorithm is compared to nine state-of-the-art approaches. All evaluations are conducted using the protocols and codes provided by the superpixel benchmark library \citep{stutz2018superpixels}. The algorithms for SLIC \citep{achanta2012slic}, SEEDS \citep{van2012seeds}, LSC \citep{li2015superpixel} are provided by the OpenCV library, the methods of ETPS \citep{yao2015real}, ERS \citep{liu2011entropy} are provided in superpixel benchmark \citep{stutz2018superpixels}, the methods of SSN \citep{jampani2018superpixel}, SPFCN \citep{yang2020superpixel}, LNSNET \citep{zhu2021learning} and SICLE \citep{belem2022efficient} are implemented from the corresponding authors. The evaluation metrics for superpixel include the achievement segmentation accuracy (ASA) and boundary recall and precision (BR-BP). ASA quantifies the achievable accuracy for segmentation using the superpixels as preprocessing step, BR and BP measure the boundary adherence of superpixels given the ground truth. 

\begin{figure}[H]
	\centering
	\includegraphics[width=\textwidth]{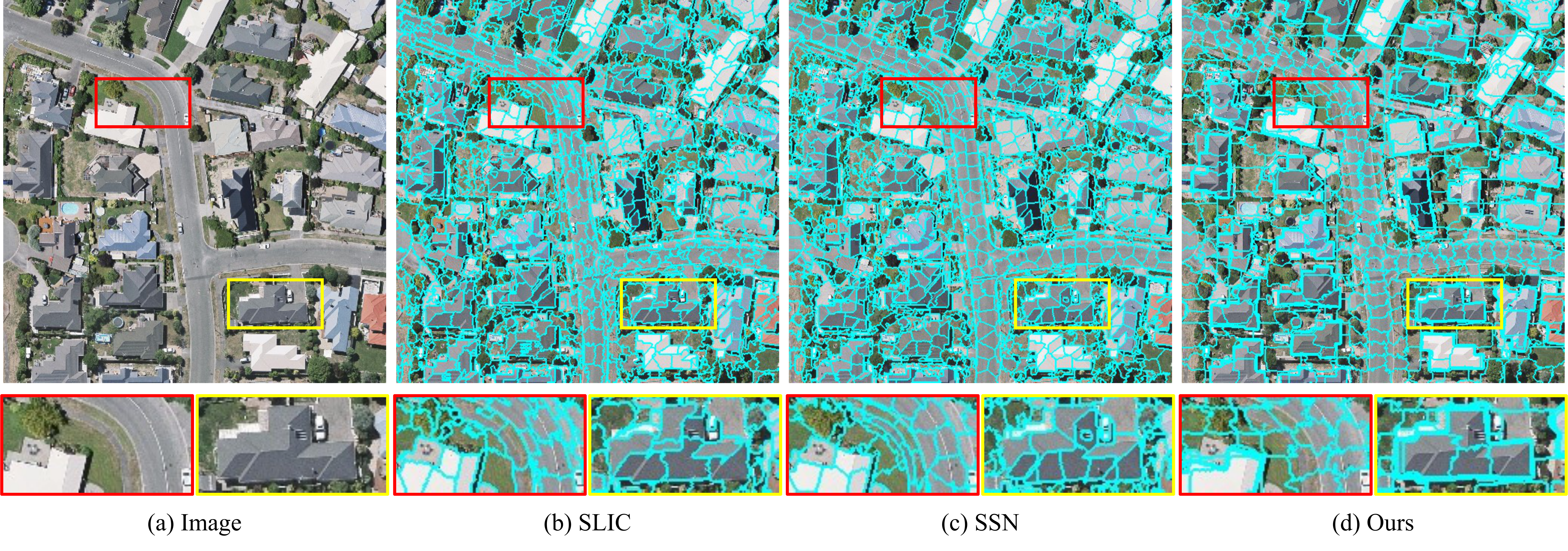}
	\caption{A comparative analysis of superpixel segmentation outcomes is conducted on the WHU dataset.}
	\label{fig:spWHU}
\end{figure}

\begin{figure}[H]
	\centering
	\includegraphics[width=\textwidth]{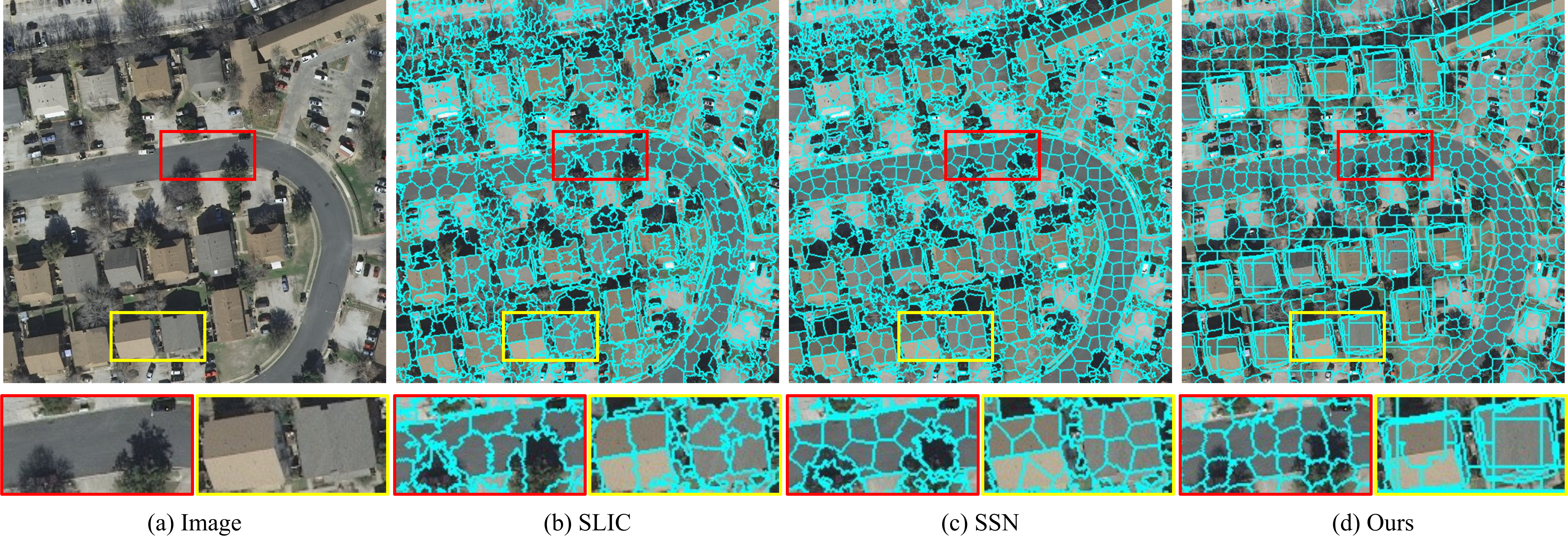}
	\caption{A comparative analysis of superpixel segmentation outcomes is conducted on the INRIA dataset.}
	\label{fig:spInria}
\end{figure}

\begin{figure}[H]
	\centering
	\includegraphics[width=\textwidth]{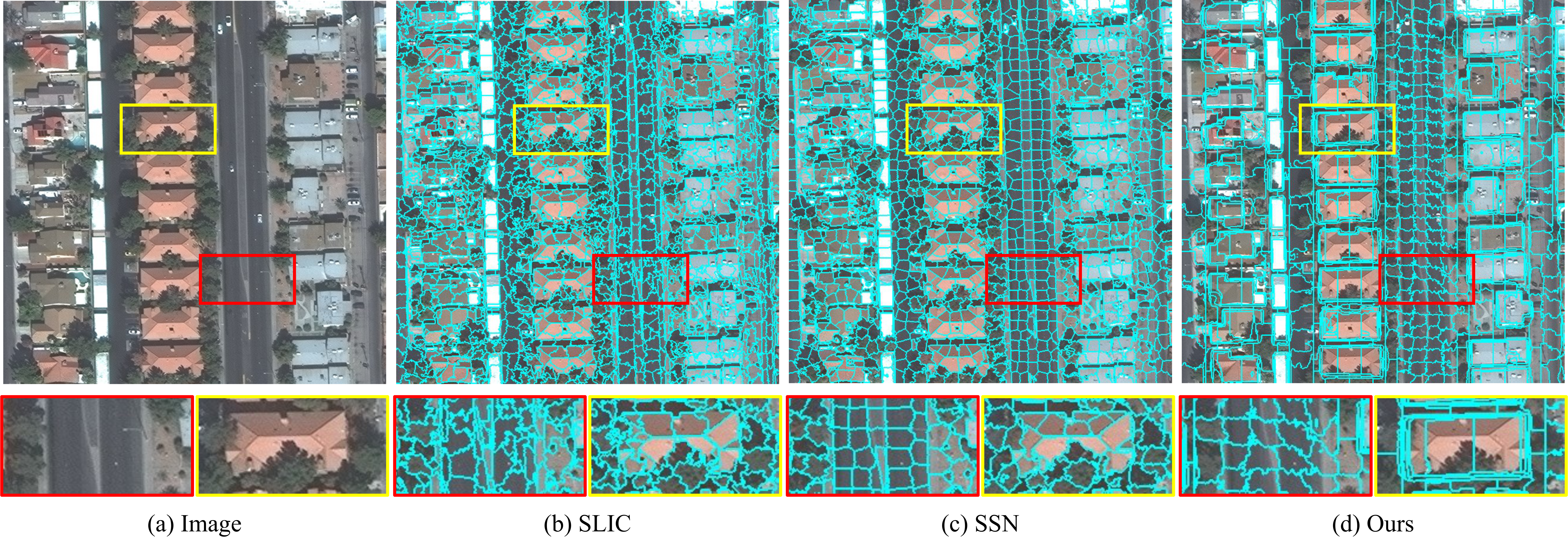}
	\caption{A comparative analysis of superpixel segmentation outcomes is conducted on the Vegas dataset.}
	\label{fig:spVegas}
\end{figure}

For qualitative results, Fig. \ref{fig:spWHU}, \ref{fig:spInria} and \ref{fig:spVegas} present the details of the superpixel segmentation outcomes and compares them to two typical approaches, such as SLIC with classical clustering methods and SSN for learned methods. It is evident that the proposed method generates superpixels that form an approximately uniform grid in the background region, while the edges of superpixels in the building region exhibit strong consistency with the building edge orientation. This demonstrates that the proposed method is exclusively sensitive to building edges, resulting in superior boundary segmentation precision. In contrast, other methods cause superpixels to generally respond to various differences in images, rendering them insensitive to buildings and leading to inaccurate and incomplete building boundaries.

\begin{figure}[H]
	\centering
	\begin{subfigure}{0.32\textwidth}
		\includegraphics[width= \textwidth]{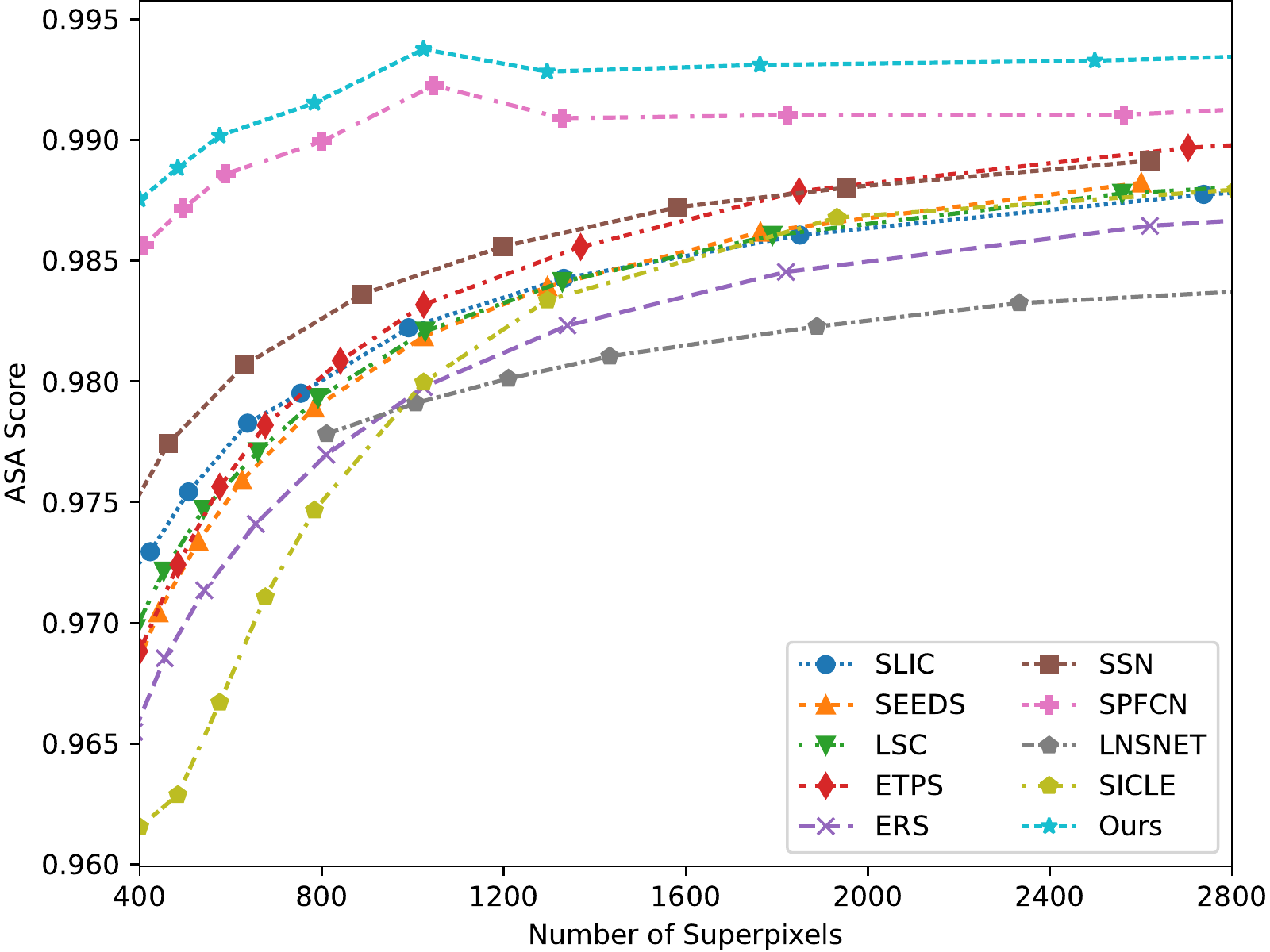}
	\end{subfigure}
	\begin{subfigure}{0.32\textwidth}
		\includegraphics[width= \textwidth]{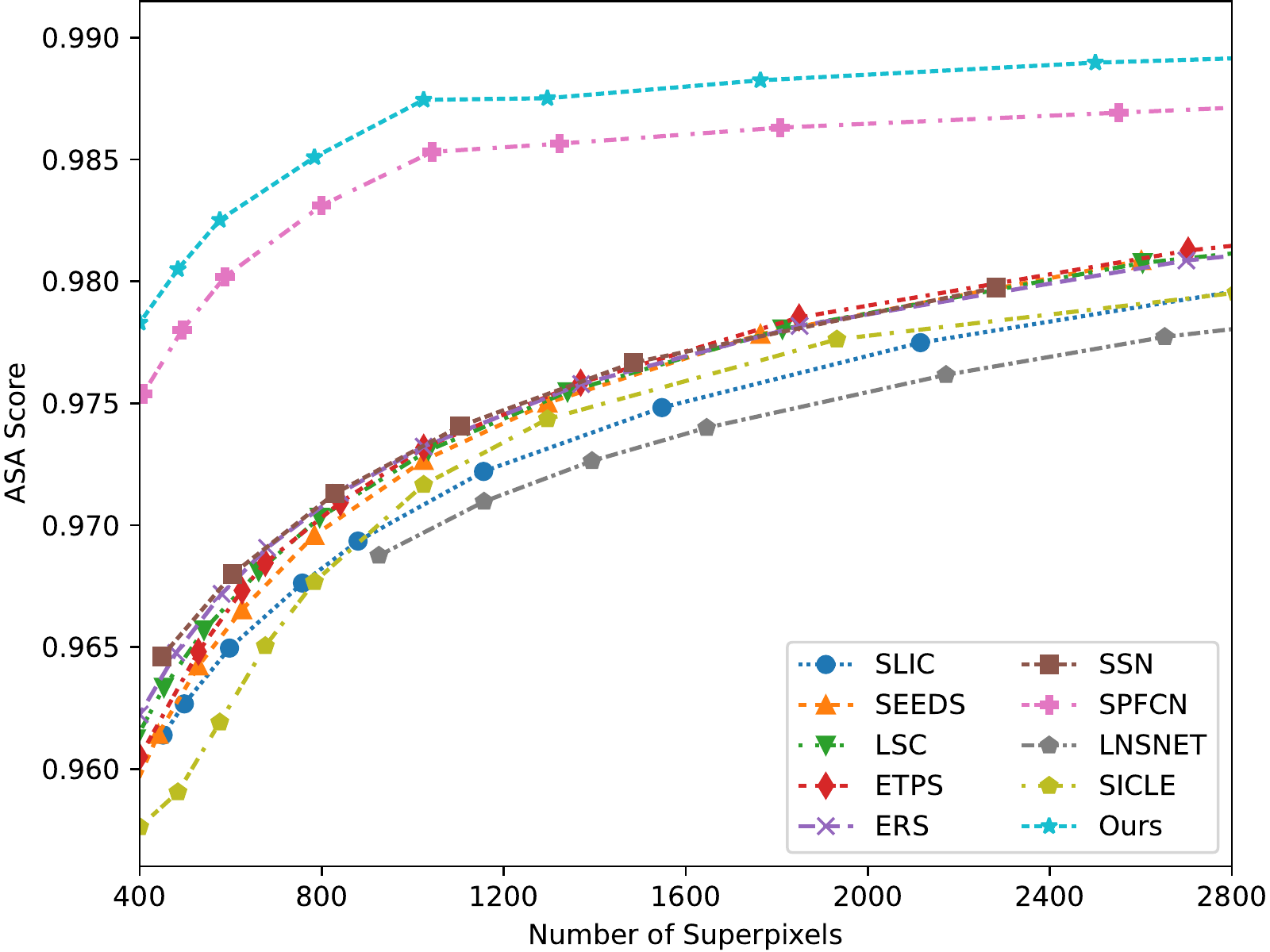}
	\end{subfigure}
	\begin{subfigure}{0.32\textwidth}
		\includegraphics[width= \textwidth]{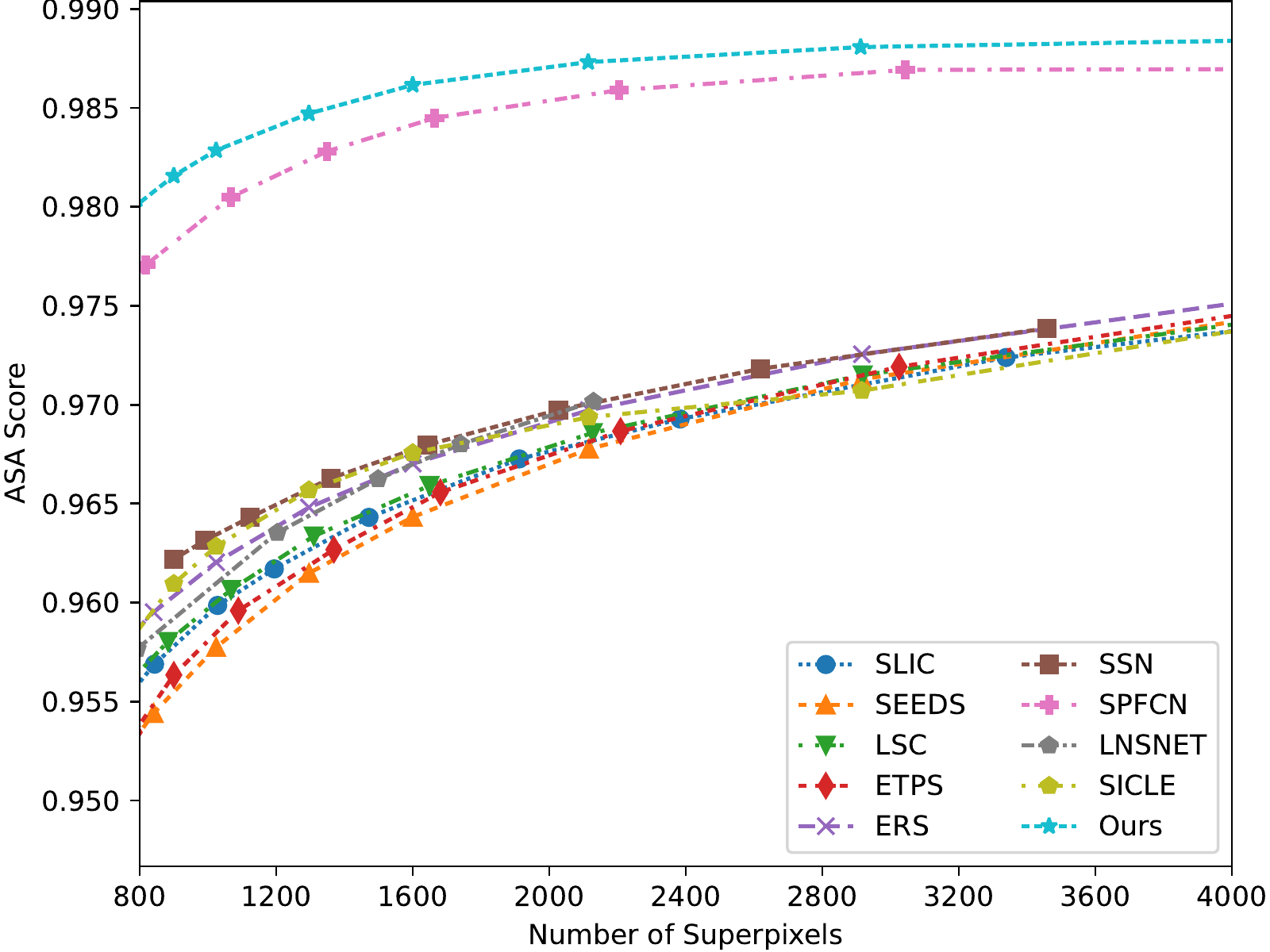}
	\end{subfigure}

	\begin{subfigure}{0.32\textwidth}
		\includegraphics[width= \textwidth]{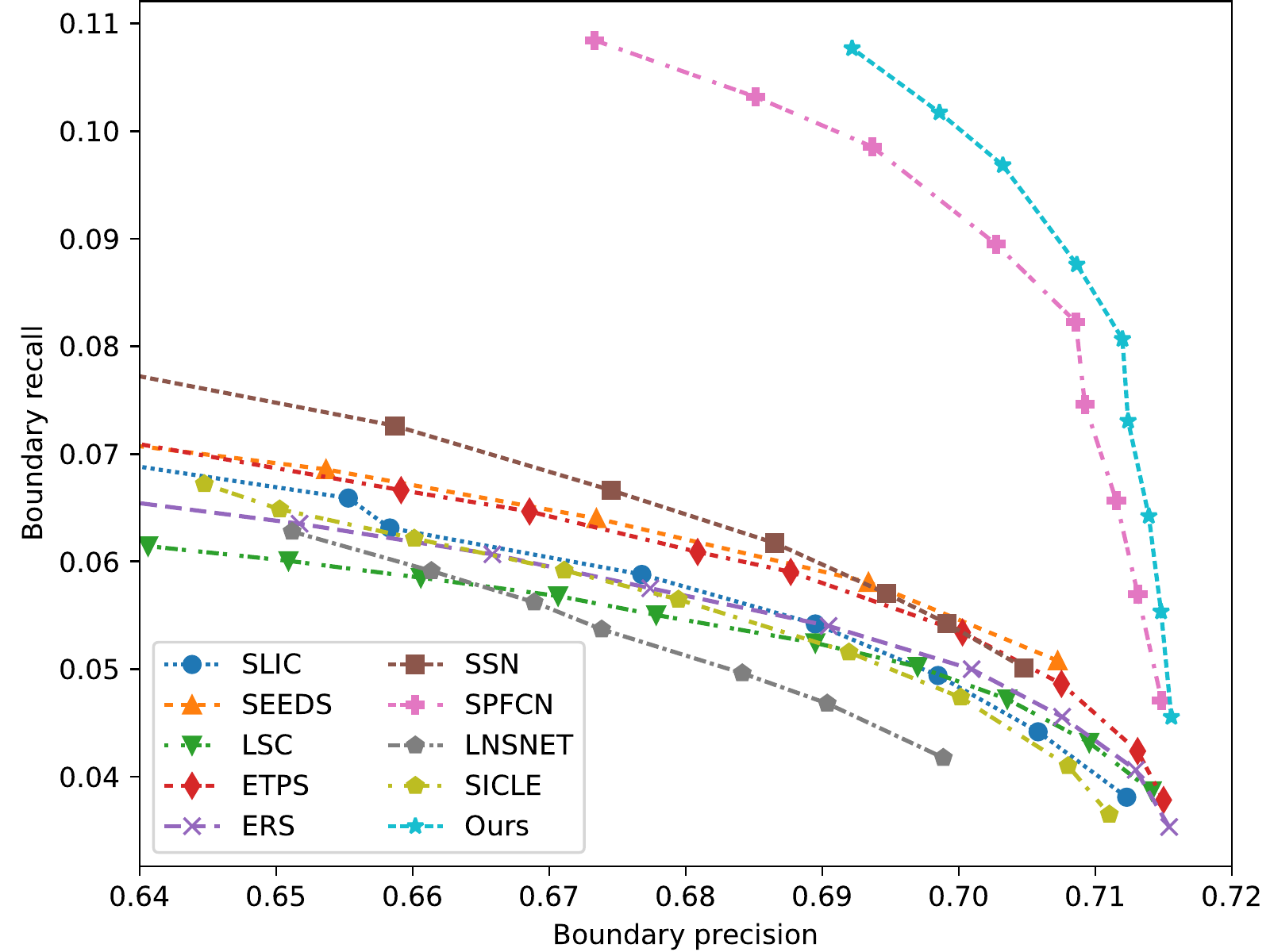}
		\caption{WHU}
	\end{subfigure}
	\begin{subfigure}{0.32\textwidth}
		\includegraphics[width= \textwidth]{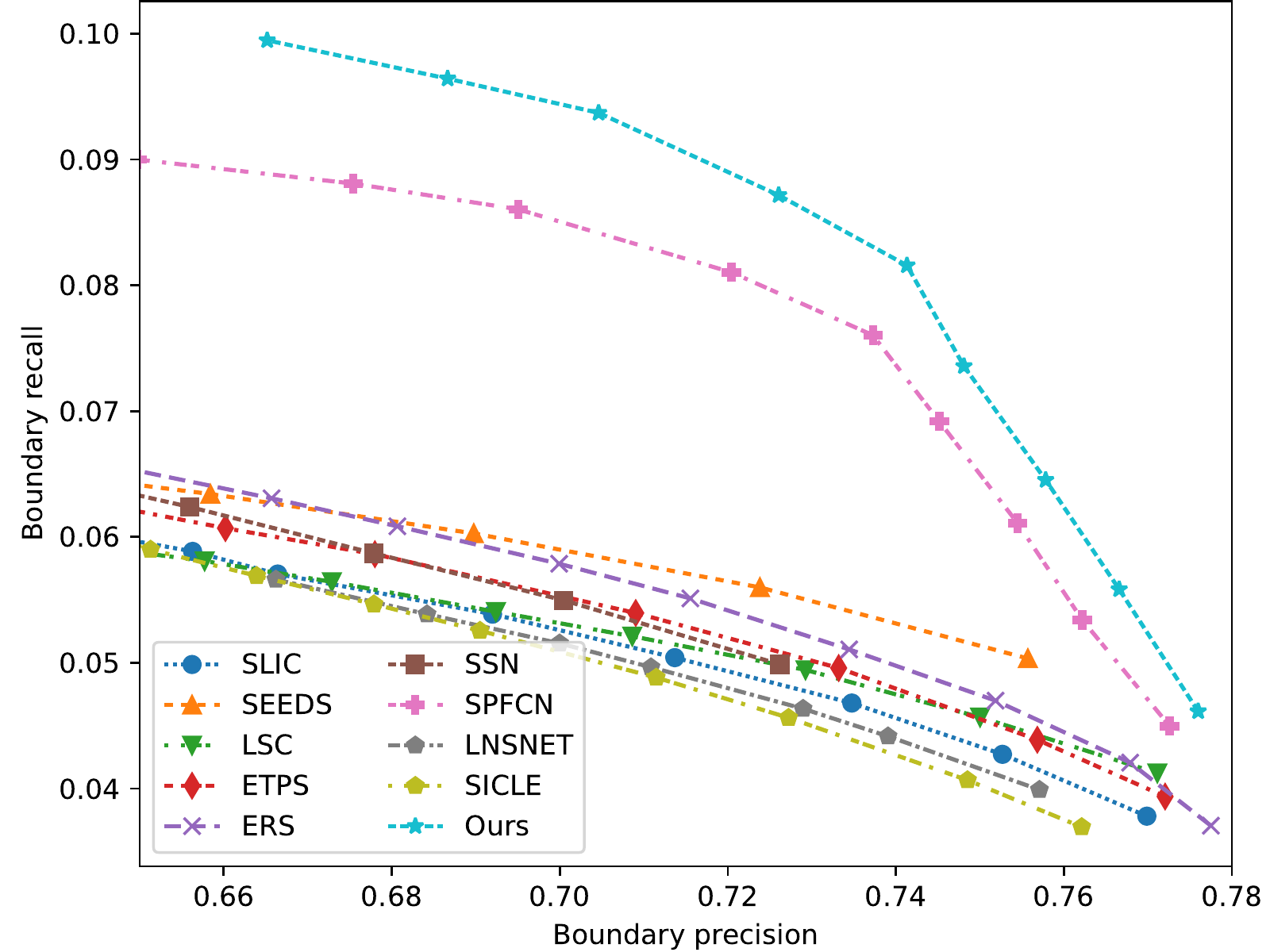}
		\caption{Inria}
	\end{subfigure}
	\begin{subfigure}{0.32\textwidth}
		\includegraphics[width= \textwidth]{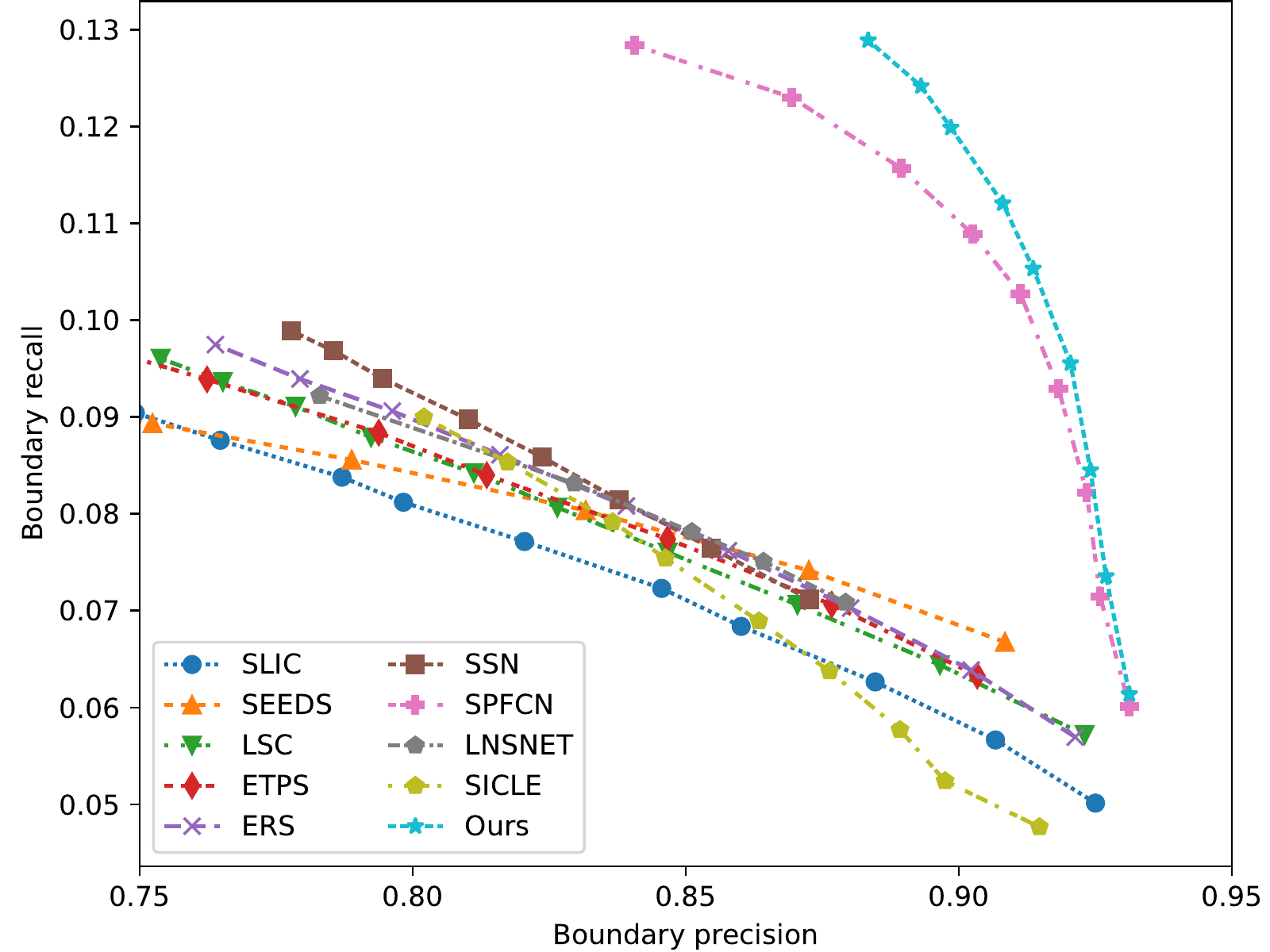}
		\caption{Vegas}
	\end{subfigure}

	\caption{A quantitative comparison of the superpixel segmentation results. It should be noted that SPFCN is solely trained for the superpixel task, while the proposed method also leverages the learned features for the subsequent segmentation task.}
	\label{fig:evl1}
\end{figure}

\begin{table}[b]
	\caption{A comparison of superpixel metrics with approximately 1000 superpixels is presented. Bold symbols indicate the most favorable results.}
	\centering
	\begin{tabular}{lrrrrrr}
		\hline
		\multicolumn{1}{l}{\multirow{2}{*}{Method}} & \multicolumn{2}{c}{WHU} & \multicolumn{2}{c}{Inria} & \multicolumn{2}{c}{Vegas} \\ \cline{2-7}
		\multicolumn{1}{l}{}                        & ASA(\%)  & BR(\%)       & ASA(\%)  & BR(\%)         & ASA(\%)  & BR(\%)         \\ \hline
		ERS    \citep{liu2011entropy}               & 97.98    & 69.05        & 97.32    & 71.55          & 96.20    & 81.60          \\
		SLIC   \citep{achanta2012slic}              & 98.22    & 67.68        & 96.93    & 66.64          & 95.99    & 78.70          \\
		SEEDS  \citep{van2012seeds}                 & 98.19    & 65.36        & 97.27    & 65.84          & 95.77    & 70.69          \\
		LSC    \citep{li2015superpixel}             & 98.21    & 68.95        & 97.31    & 69.24          & 96.06    & 79.24          \\
		ETPS   \citep{yao2015real}                  & 98.32    & 68.77        & 97.32    & 67.80          & 95.96    & 76.23          \\
		SICLE  \citep{belem2022efficient}           & 98.00    & 69.20        & 97.17    & 71.15          & 96.28    & 84.63          \\ \hline
		SSN    \citep{jampani2018superpixel}        & 98.36    & 67.45        & 97.41    & 67.78          & 96.31    & 78.55          \\
		SPFCN  \citep{yang2020superpixel}           & 99.23    & 70.86        & 98.53    & 73.73          & 98.05    & 88.95          \\
		LNSNET \citep{zhu2021learning}              & 97.91    & 66.14        & 96.88    & 66.62          & 96.35    & 82.96          \\
		SuperpixelGraph                             & \textbf{99.38} & \textbf{71.20} & \textbf{98.74} & \textbf{74.13} & \textbf{98.29} & \textbf{89.85} \\ \hline
	\end{tabular}
    \label{tab:spCompare} 
\end{table}

For the quantitative assessments, we classify the techniques into two primary categories: non-learned methods (top rows of Table \ref{tab:spCompare}) and learned methods (bottom rows of Table \ref{tab:spCompare}). These evaluations focus on the delineation of ground truth building polygons. Notably, our proposed methodology exhibits a remarkable improvement in performance, surpassing existing approaches by a substantial margin when analyzing building boundaries (Fig. \ref{fig:evl1}). We additionally present the standard performance metrics while maintaining the number of superpixels at approximately 1000, demonstrating consistent results. Furthermore, it is imperative to emphasize that the learned methods (bottom rows) undergo fine-tuning utilizing the same segmentation map containing buildings as the proposed methods. This process accentuates the significance of semantic sensitivity in achieving exceptional performance. The enhanced preservation of building contours will substantially augment subsequent building extraction endeavors.

\subsection{Evaluation of Building Extraction}
\label{s:preAndLoc}
In this section, we meticulously evaluate the performance of the proposed methodology for building extraction by examining both the pixel-level accuracy and the vector-level correspondence. Despite the primary focus of this paper being the generation of vector graphics, the pixel-level assessment remains crucial, as we employ conventional simplification and regularization techniques.

\begin{figure}[H]
	\centering
	\begin{subfigure}[b]{0.24\textwidth}
		\includegraphics[width= \textwidth]{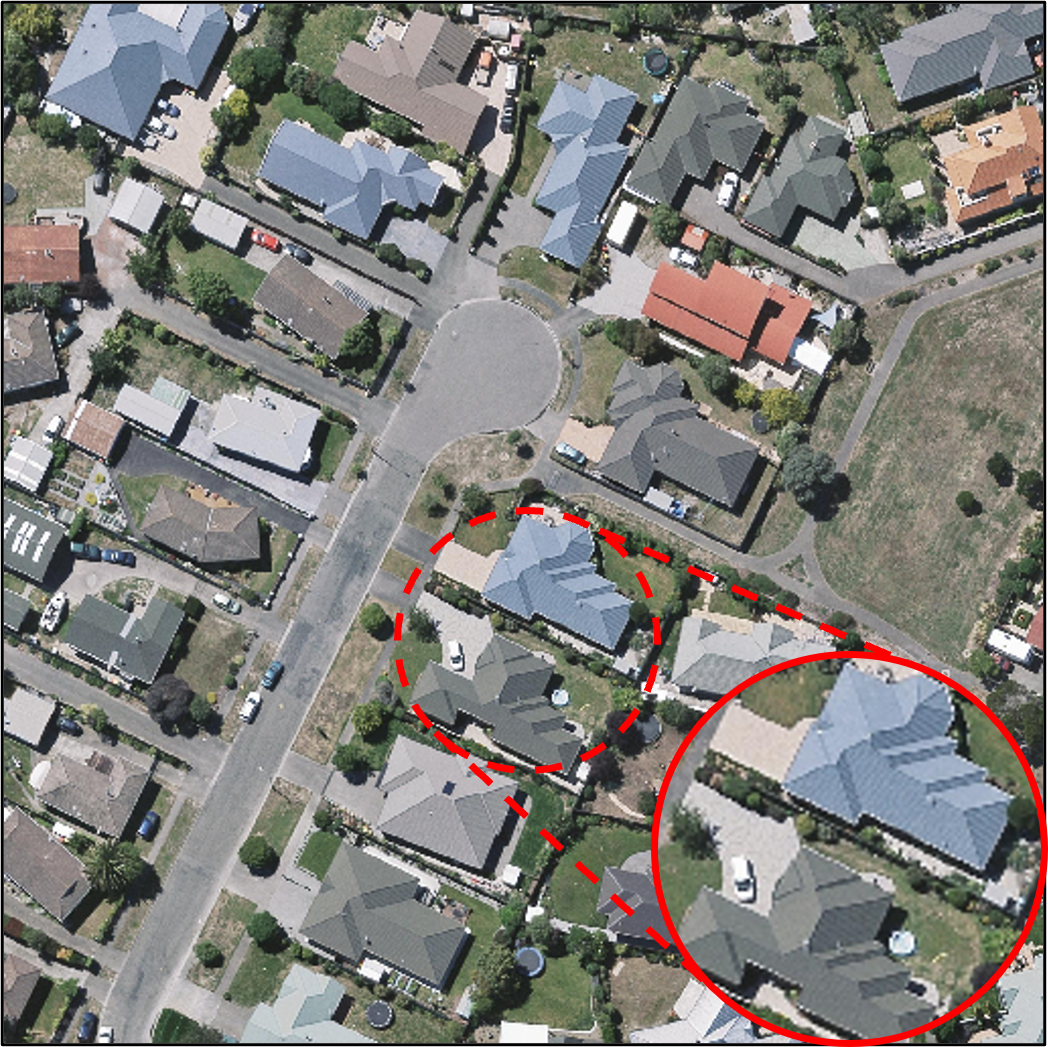}
	\end{subfigure}	
	\begin{subfigure}[b]{0.24\textwidth}
		\includegraphics[width= \textwidth]{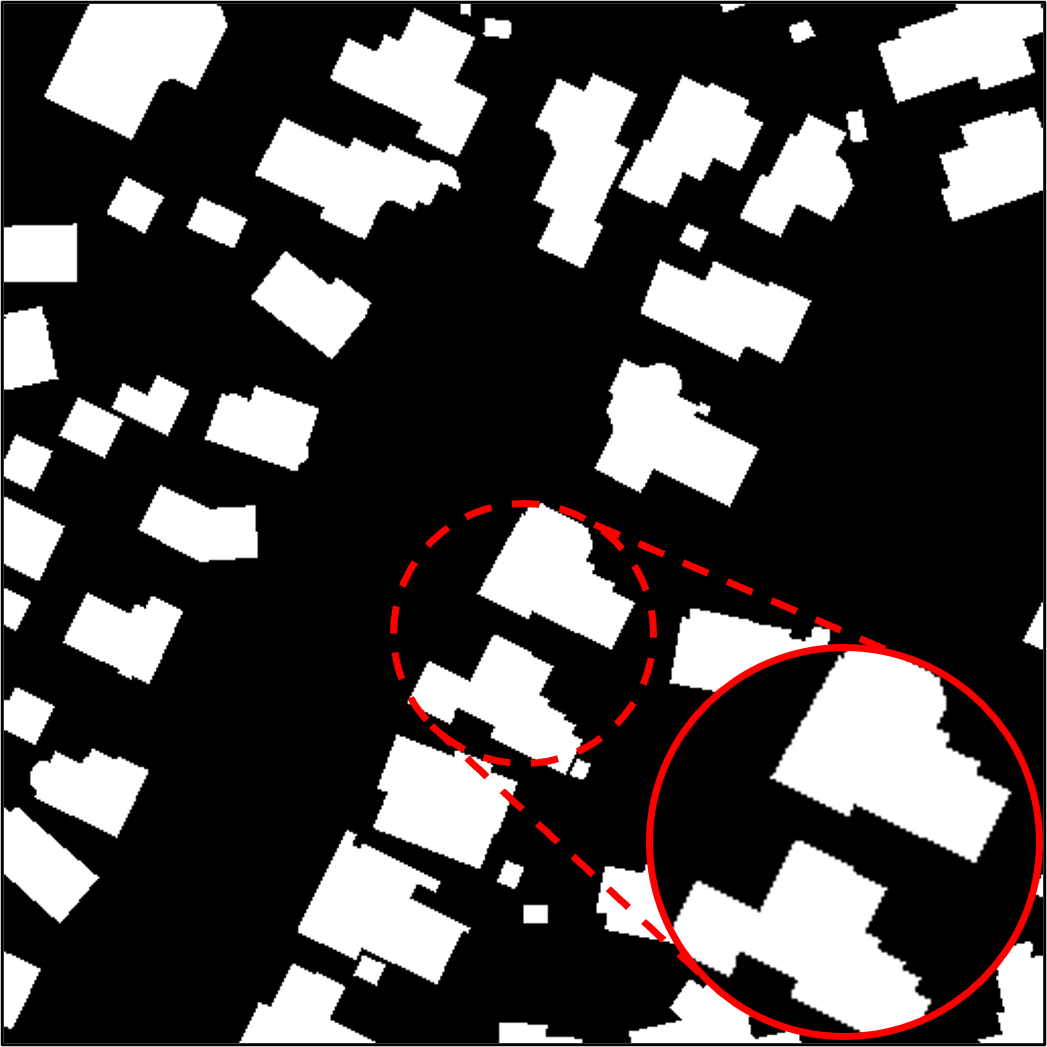}
	\end{subfigure}	
    \begin{subfigure}[b]{0.24\textwidth}
		\includegraphics[width= \textwidth]{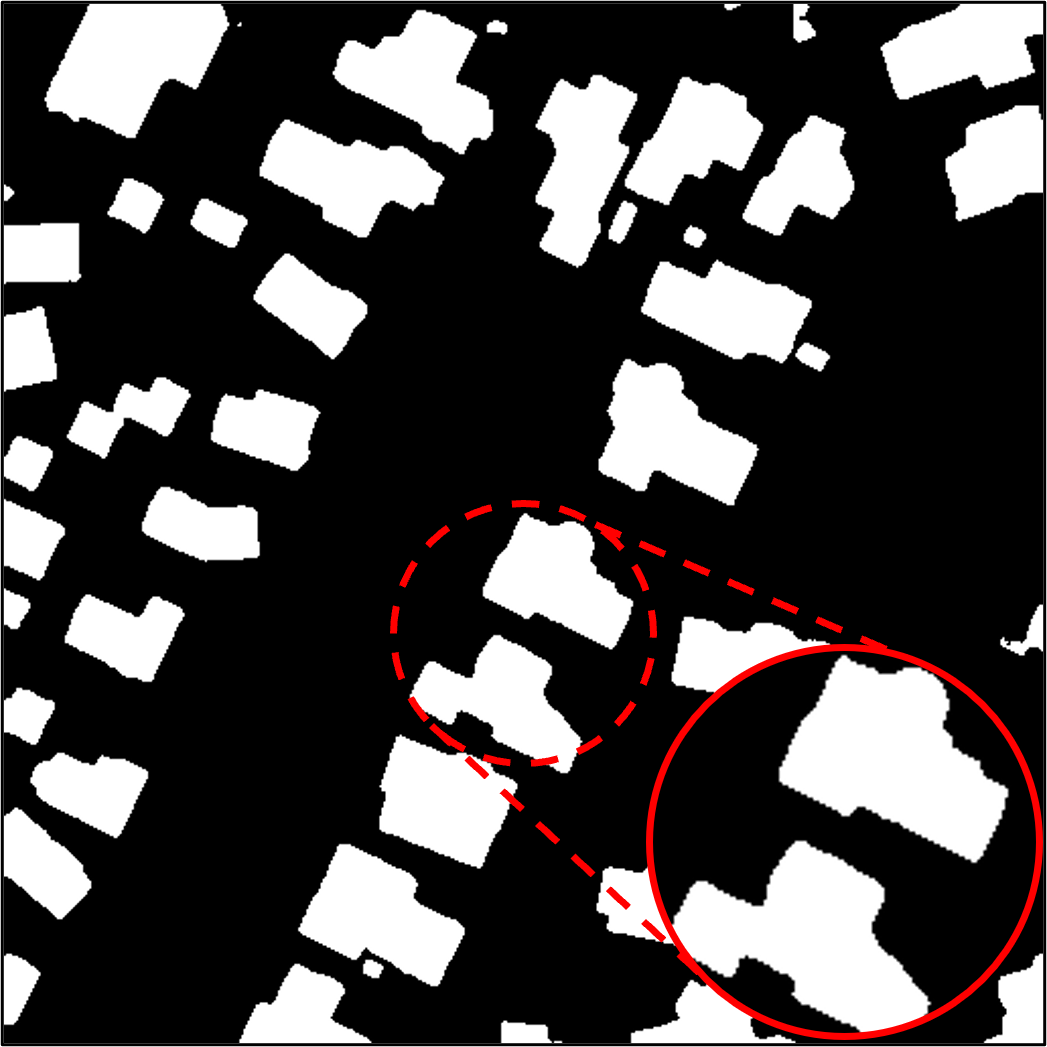}
	\end{subfigure}	
    \begin{subfigure}[b]{0.24\textwidth}
		\includegraphics[width= \textwidth]{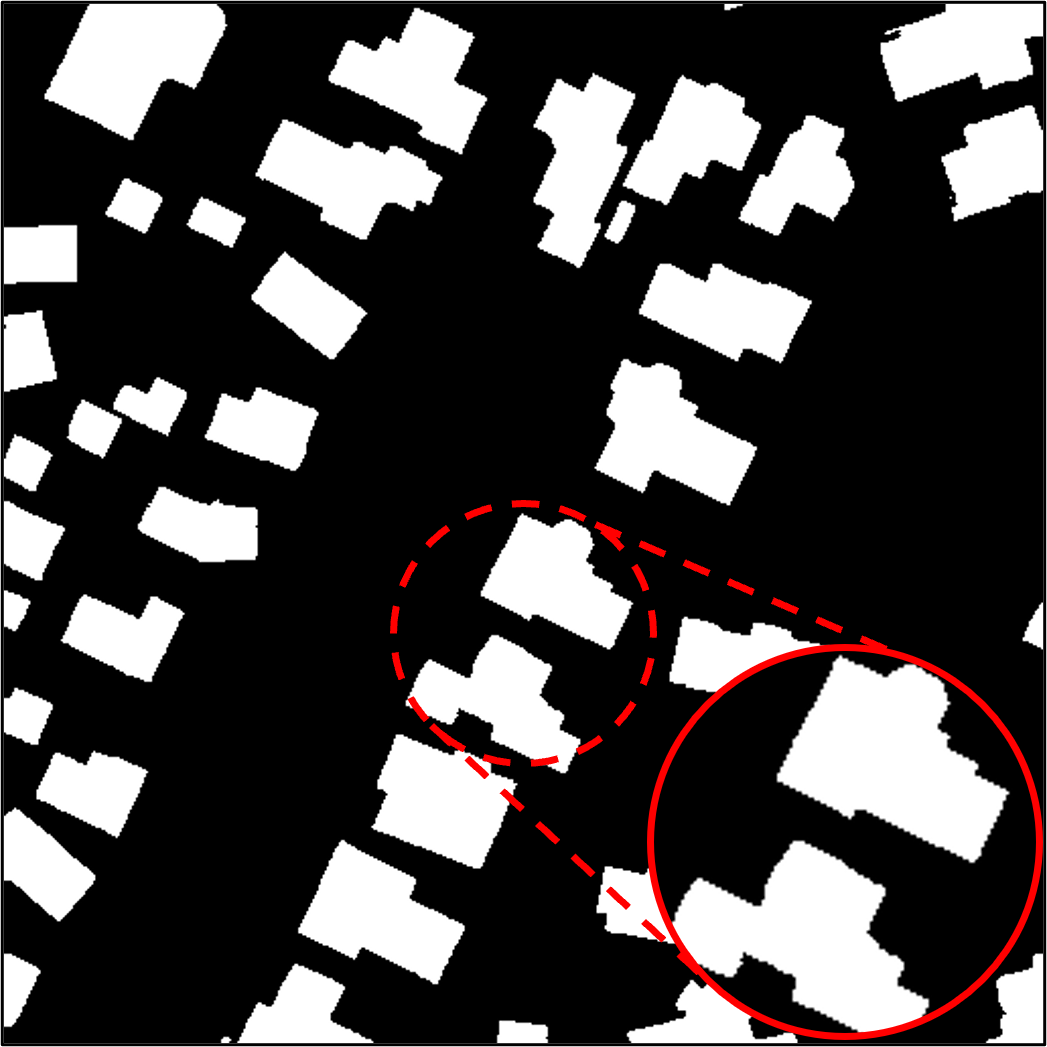}
	\end{subfigure}	
	
	\begin{subfigure}[b]{0.24\textwidth}
		\includegraphics[width= \textwidth]{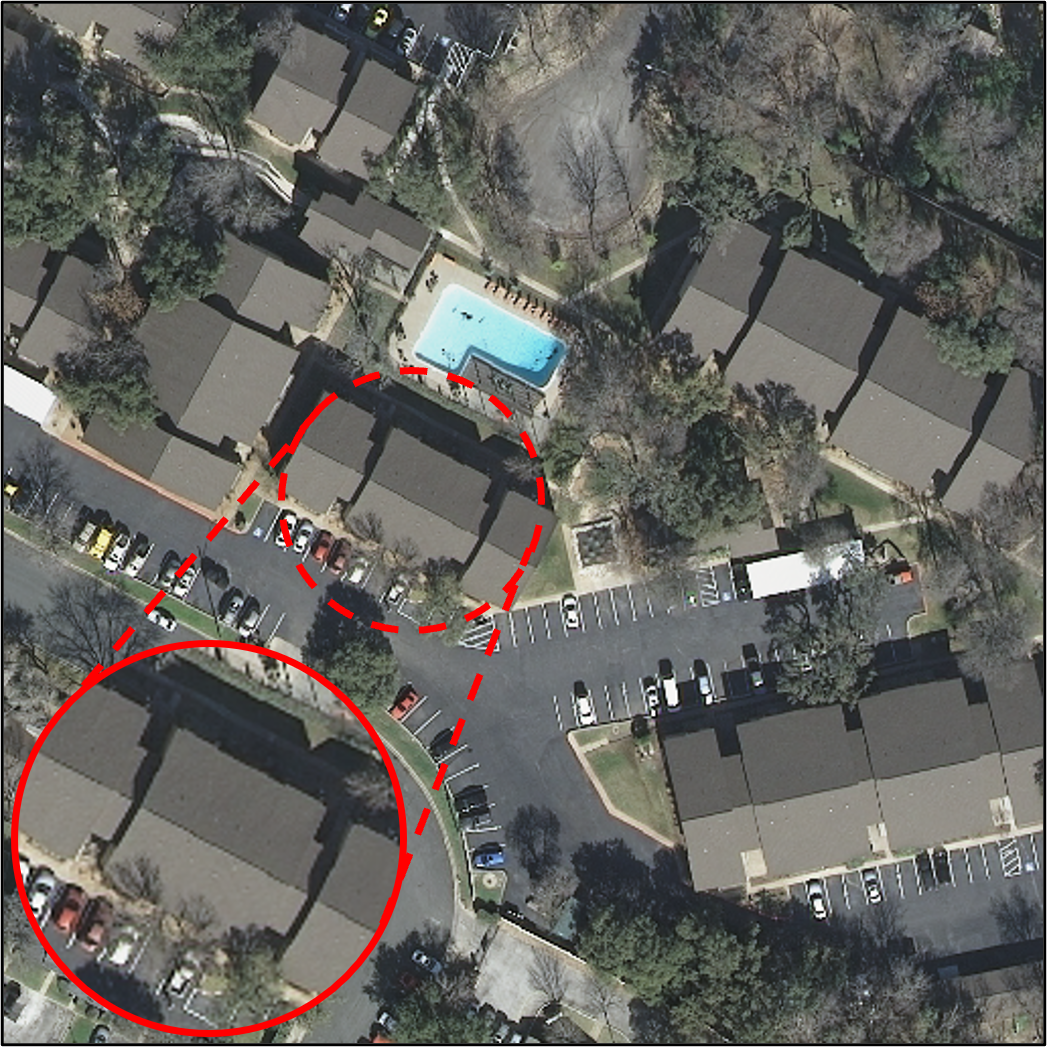}
	\end{subfigure}	
	\begin{subfigure}[b]{0.24\textwidth}
		\includegraphics[width= \textwidth]{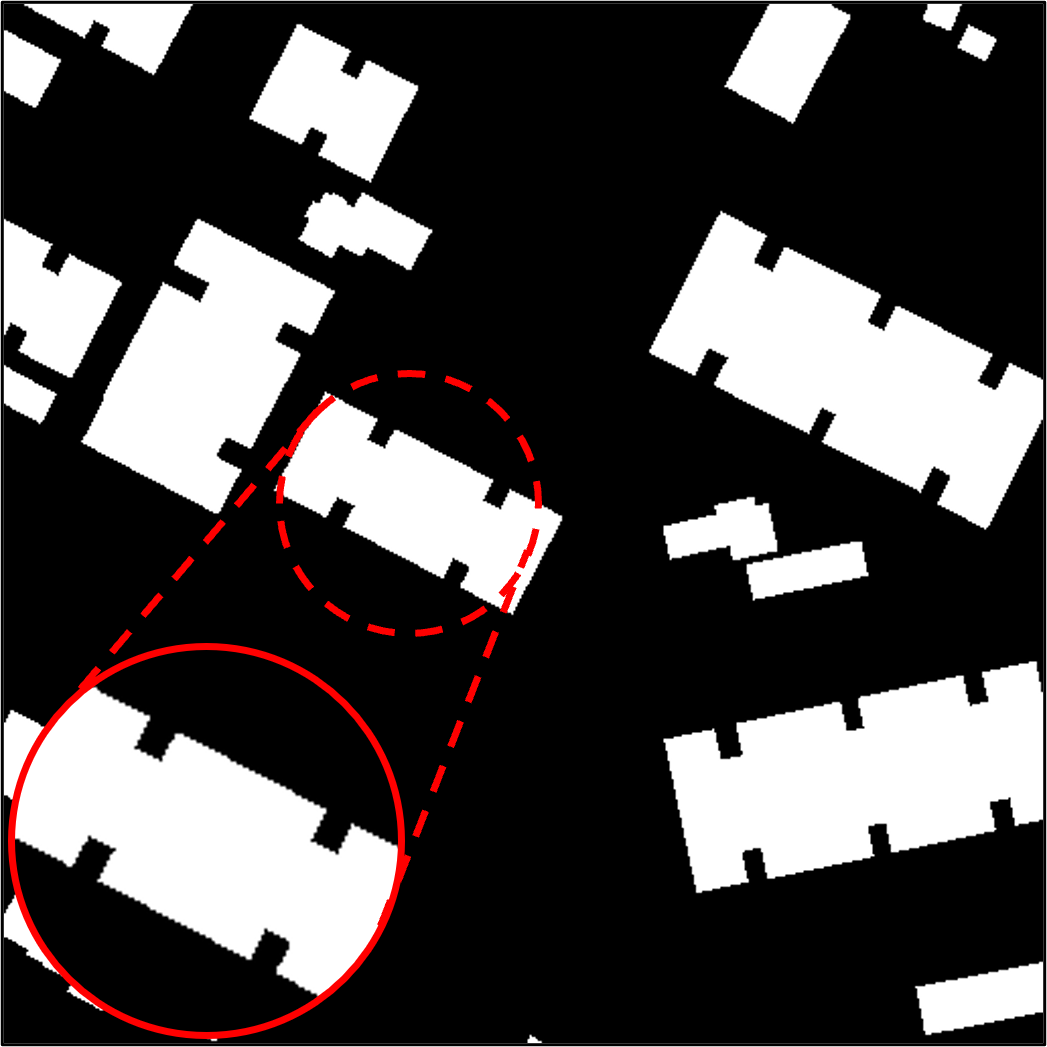}
	\end{subfigure}	
    \begin{subfigure}[b]{0.24\textwidth}
		\includegraphics[width= \textwidth]{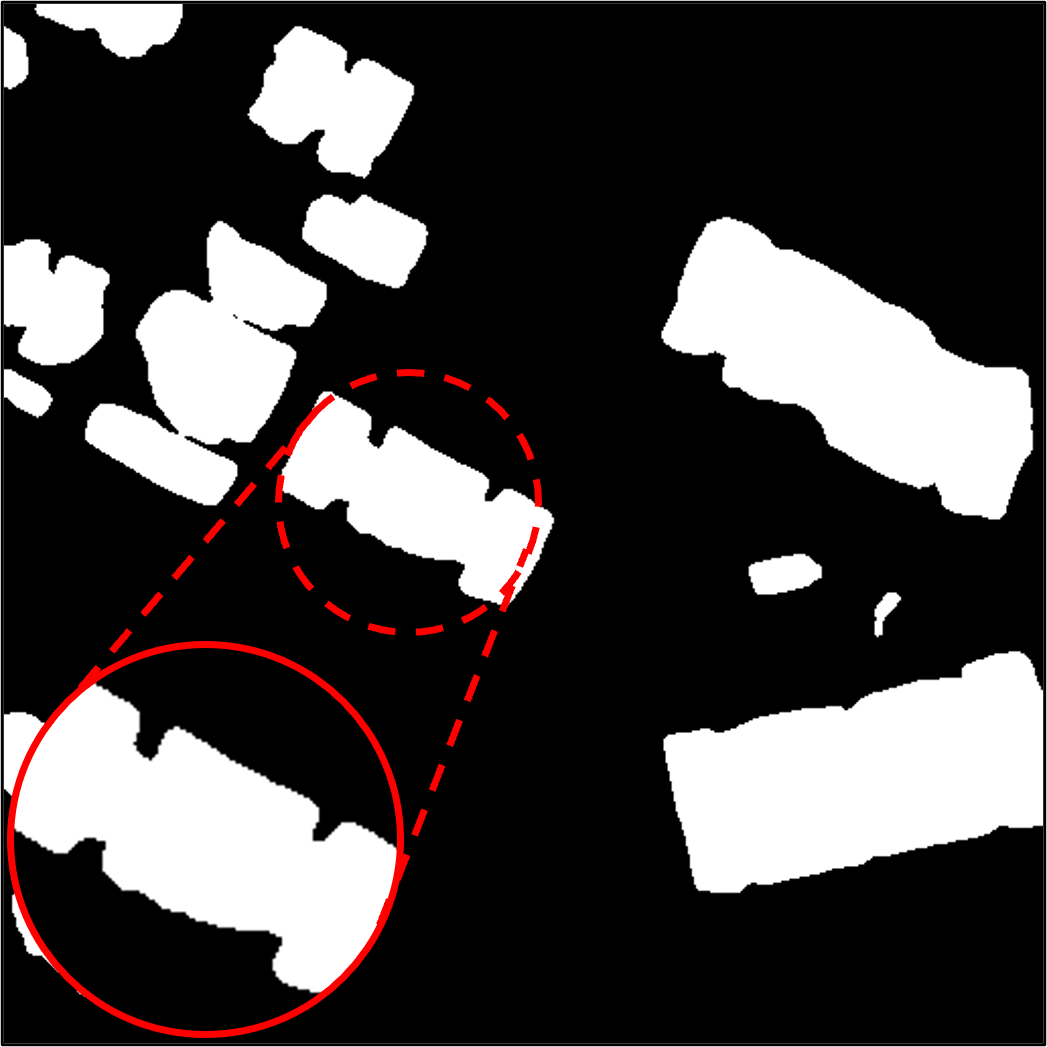}
	\end{subfigure}	
    \begin{subfigure}[b]{0.24\textwidth}
		\includegraphics[width= \textwidth]{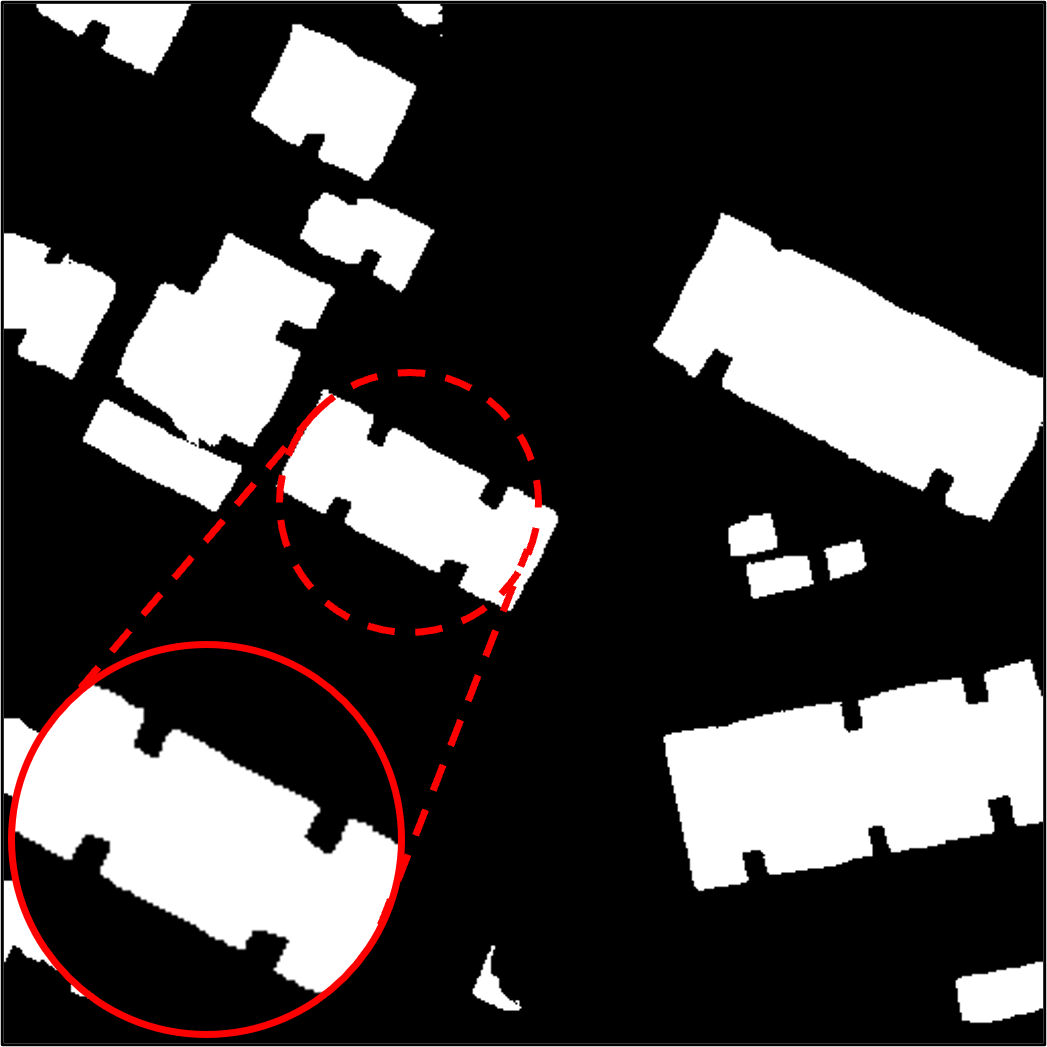}
	\end{subfigure}	

    \begin{subfigure}[b]{0.24\textwidth}
        \includegraphics[width= \textwidth]{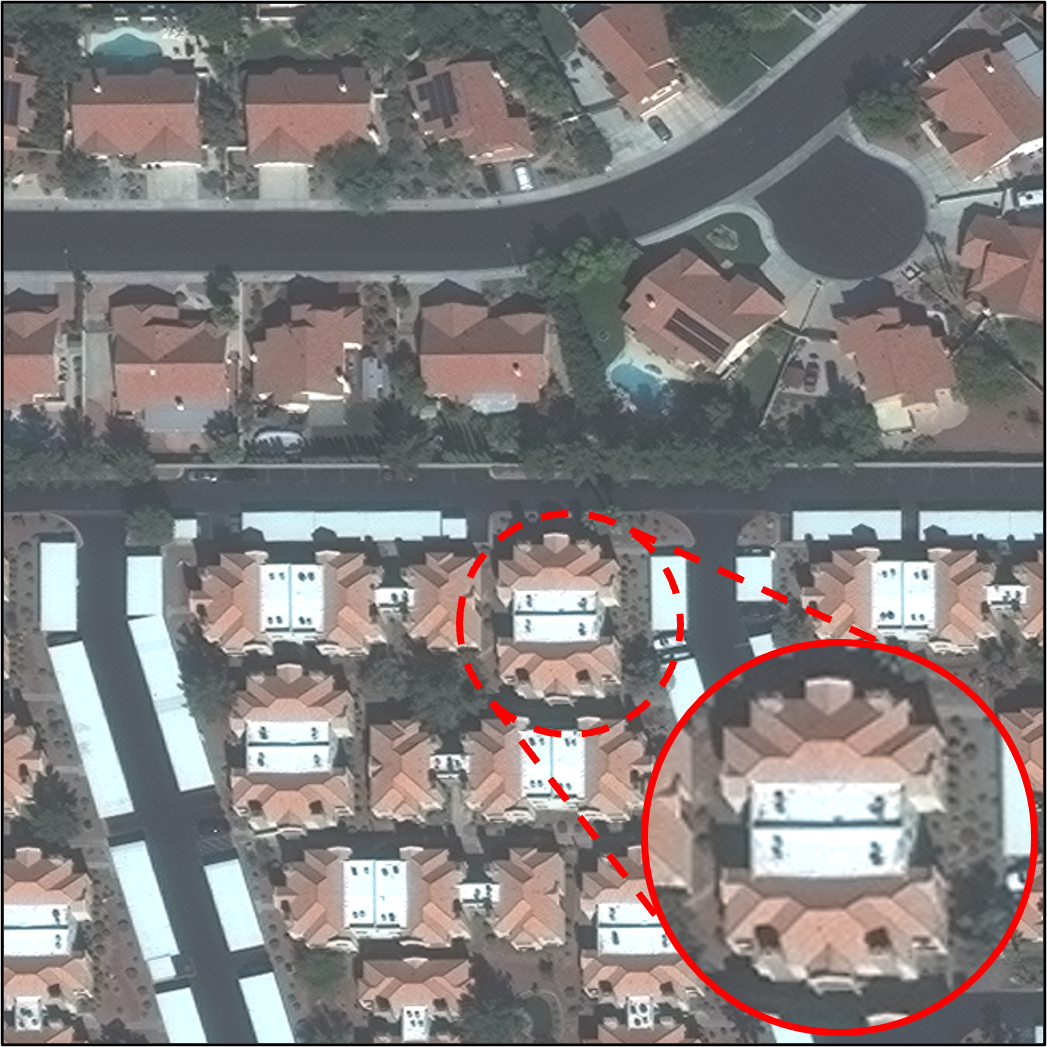}
        \caption{Image}
	\end{subfigure}	
	\begin{subfigure}[b]{0.24\textwidth}
		\includegraphics[width= \textwidth]{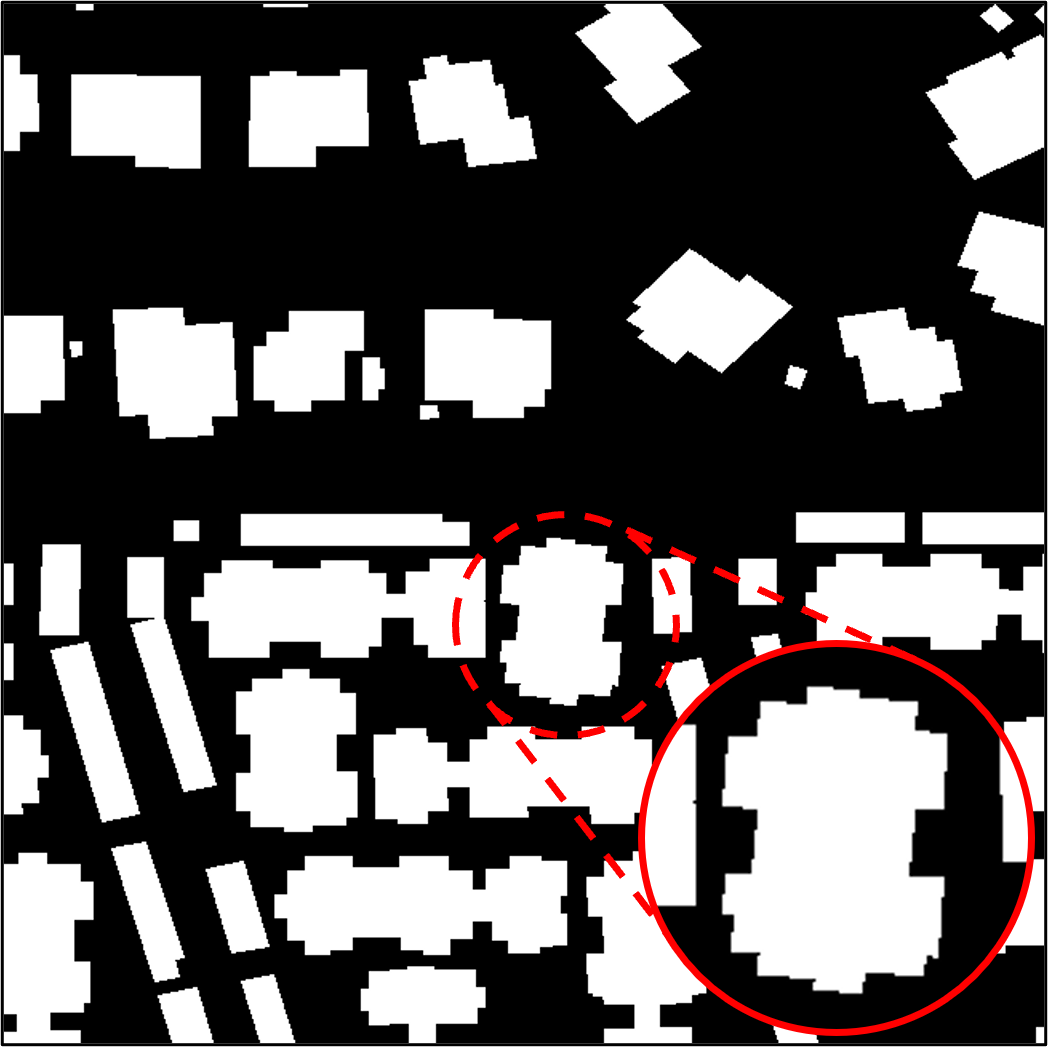}
        \caption{GT}
	\end{subfigure}	
    \begin{subfigure}[b]{0.24\textwidth}
		\includegraphics[width= \textwidth]{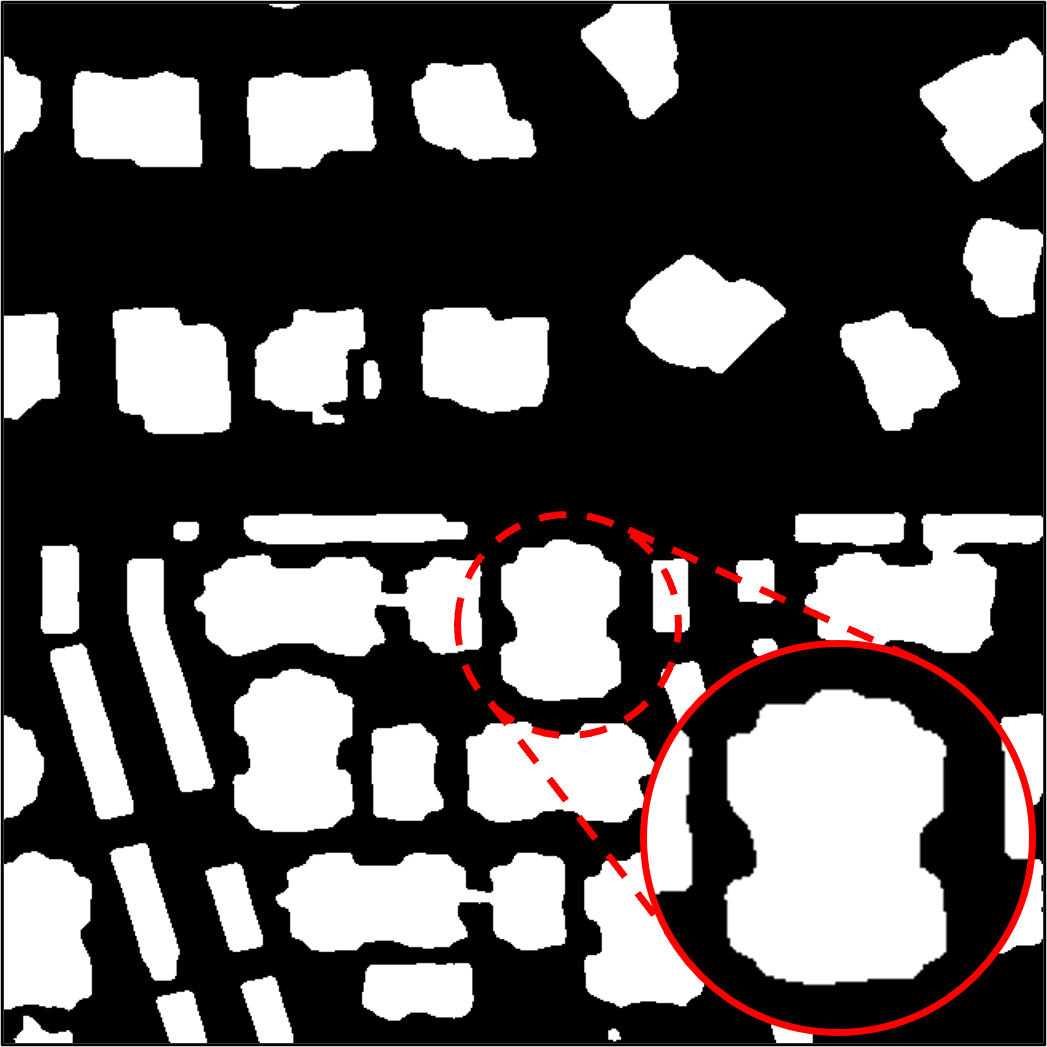}
        \caption{Baseline}
	\end{subfigure}	
    \begin{subfigure}[b]{0.24\textwidth}
		\includegraphics[width= \textwidth]{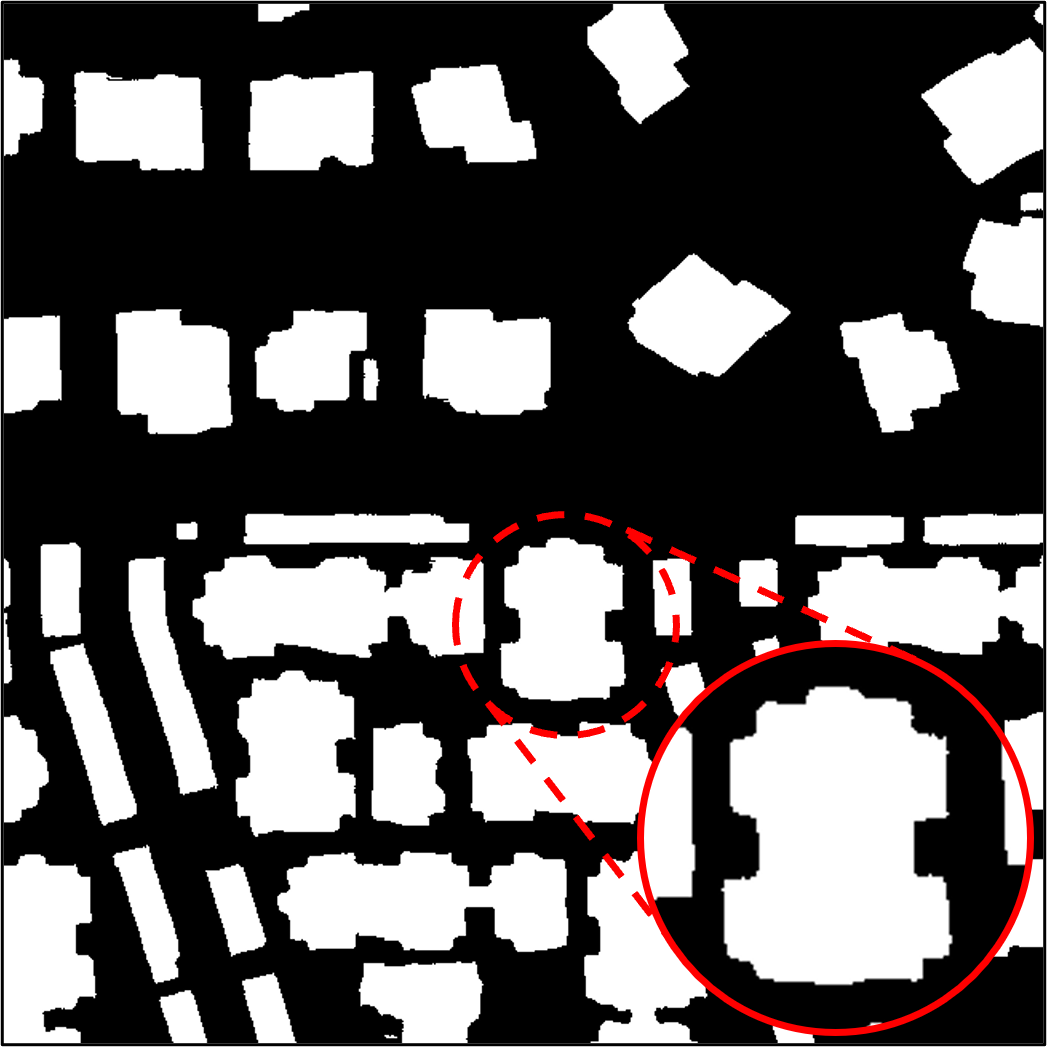}
        \caption{Ours}
	\end{subfigure}	

	\caption{Comparison of pixel-wise segmentation results. The first to the third rows demonstrates a typical tile for the WHU, Inria and Vegas datasets, respectively. The dotted circles demonstrates enlarged view of a typical building.}
    \label{fig:evl3}
\end{figure}

\textbf{Pixel-wise Segmentation}. To assess the influence of superpixels within the network on the precision of building detection, we employ the U$^2$-Net for semantic segmentation (utilizing the same encoder as our superpixel segmentation network) as the baseline, i.e., classification heads following the feature map $f$. The segmentation results are depicted in Fig. \ref{fig:evl3}. It is apparent that the SuperpixelGraph is capable of generating more distinct and sharp boundaries compared to the baseline approach. This can be attributed to the fact that traditional networks like U$^2$-Net frequently employ convolution and pooling processes, which cannot preserve boundaries in the encoder. In contrast, our method's superpixel generation network can provide boundary-preserving superpixels.

Furthermore, to compare the efficacy of the learned pixel-superpixel association matrix $Q$ (Equation \ref{eq:agg_v}), we also perform ablation analyses, as displayed in Table \ref{tab:pixCompare}. The method U$^2$-Net+SLIC refers to averaging the segmentation results using the SLIC superpixel after the baseline. U$^2$-Net+SLIC+GAT represents the results obtained with average superpixel pooling using the SLIC results, followed by appending graph optimization. The quantitative outcomes are provided in Table \ref{tab:pixCompare}, where it can be observed that the disparities between SuperpixelGraph and the baseline in terms of pixel-wise metrics are relatively minor. However, as we will demonstrate subsequently, the vector-level performance differences are substantially more pronounced. Additionally, naively substituting the graph construction with SLIC does not enhance performance or, in some cases, even converge (as seen in the Vegas dataset). This underscores the significance of the soft superpixel aggregation facilitated by the association matrix $Q$.

\begin{table}[H]
\caption{Pixel-level building detection accuracies. The bolded cells represent the best results achieved for each individual dataset.}
\centering
\begin{tabular}{clrrrr}
\hline
\multicolumn{1}{l}{Dataset} & Model                 & Precision(\%)  & Recall(\%)     & F1 score(\%)   & IoU(\%)        \\ \hline
\multirow{4}{*}{WHU}        & U$^2$-Net             & 92.97          & \textbf{94.59} & 93.77          & 88.27          \\
							& U$^2$-Net+SLIC        & 90.86          & 88.25          & 89.53          & 81.05          \\ 
							& U$^2$-Net+SLIC+GAT    & 90.18          & 89.29          & 89.73          & 81.38          \\
                            & SuperpixelGraph       & \textbf{93.94} & 94.45          & \textbf{94.19} & \textbf{89.02} \\ \hline
\multirow{4}{*}{Inria}      & U$^2$-Net             & \textbf{87.70} & 86.12          & \textbf{86.90} & \textbf{76.84} \\
							& U$^2$-Net+SLIC        & 83.14          & 83.92          & 83.53          & 71.71          \\
							& U$^2$-Net+SLIC+GAT    & 83.75          & 80.08          & 81.87          & 69.31          \\
                            & SuperpixelGraph       & 84.76          & \textbf{88.78} & 86.73          & 76.56          \\ \hline
\multirow{4}{*}{Vegas}      & U$^2$-Net             & 92.17          & 90.91          & 91.54          & 84.39          \\
                            & U$^2$-Net+SLIC        & 88.84          & 88.00          & 88.42          & 79.24          \\
							& U$^2$-Net+SLIC+GAT    & -              & -              & -              & -              \\
                            & SuperpixelGraph       & \textbf{92.37} & \textbf{91.05} & \textbf{91.70} & \textbf{84.68} \\ \hline
\end{tabular}
\label{tab:pixCompare}  
\end{table}

\textbf{Vector-Level Segmentation Analysis}. In this section, we assess and contrast the performance of the resulting building vector polygons with existing methodologies. For U$^2$-Net \citep{qin2020u2}, which solely provides building labels, we employ the ASIP \citep{li2020approximating} and ArcGIS \citep{gribov2019optimal} techniques for the vectorization process. Furthermore, we also compare our approach to state-of-the-art end-to-end networks such as PolyWorld \citep{zorzi2022polyworld} and Frame Field Learning (FrameField) \citep{girard2021polygonal}. The evaluation metrics employed are consistent with those used by \cite{huang2021oec} and include weighted coverage (WC), boundary F-score (BF), Hausdorff distance (HD), vertex number error (VNE), and average precision (AP). The evaluation results, based on vectors that possess a minimum of 50\% intersection of union with the ground truth instances, are presented in Table \ref{tab:vecCompare}. It is important to note that the results for PolyWorld and FrameField are derived using their respective pre-trained weights, as the training codes have been retained as proprietary by the authors.

\begin{table}[H]
\caption{Vector-level building detection accuracies for the WHU Datasets. The $\uparrow$ symbol indicates that a higher value is preferable, while the $\downarrow$ symbol signifies that a lower value is more desirable. The top rows demonstrate results with an end-to-end approach, while the bottom rows showcase those using raster-to-vector conversion.}
\centering
\begin{tabular}{lrrrrrr}
\hline
\multicolumn{1}{c}{Model} & AP50(\%)$\uparrow$ & AP75(\%)$\uparrow$ & WC(\%)$\uparrow$ & BF(\%)$\uparrow$ & HD$\downarrow$ & VNE$\downarrow$ \\ \hline
PolyWorld                 & 37.34              & 23.81              & 85.04            & 20.27            & 21.60          & 3.50            \\
FrameField                & 73.50              & 54.81              & 87.28            & 22.69            & 12.78          & 2.76            \\ \hline
U$^2$-Net+ASIP            & 52.91              & 44.54              & \textbf{91.52}   & \textbf{26.05}   & 14.07          & 4.46            \\
U$^2$-Net+ArcGIS          & 82.78              & 66.10              & 91.02            & 23.65            & 13.59          & 3.14            \\
SuperpixelGraph           & \textbf{90.62}     & \textbf{78.46}     & 90.39            & 25.70            & \textbf{9.43}  & \textbf{2.40}   \\ \hline
\end{tabular}
\label{tab:vecCompare}
\end{table}

Table \ref{tab:vecCompare} presents a comparison of the vector-level metrics. The top rows display results obtained using end-to-end approaches, while the bottom rows require raster-to-vector conversion.
From the results, it is evident that U$^2$-Net+ASIP yields high WC and BF values but a low AP50 value. This can be attributed to the fact that WC and BF metrics only compute the precision of correct instances, meaning that only a significantly lower number of high-quality buildings are produced by these methods. Given that the U$^2$-Net baseline achieves almost identical pixel-wise metrics as SuperpixelGraph, it is intriguing to observe the disparity between pixel-level and vector-level scores.
It is worth noting that FrameField demonstrates a substantially better generalization capability compared to PolyWorld. This is likely due to the learning of low-level field representation, while the vector polygons are still generated using traditional tracing techniques.
Regarding U$^2$-Net+ArcGIS, the segmentation accuracy surpasses that of both U$^2$-Net+ASIP, underscoring the significance of building tracing methods for raster-to-vector conversion.
Nonetheless, the proposed methods, despite employing a conventional polygon simplification and regularization approach, still outperform the aforementioned techniques.

\subsection{Interactive editing with global optimization}
\begin{figure}[H]
	\centering
	\begin{subfigure}[b]{0.8\linewidth}
		\includegraphics[width=\linewidth]{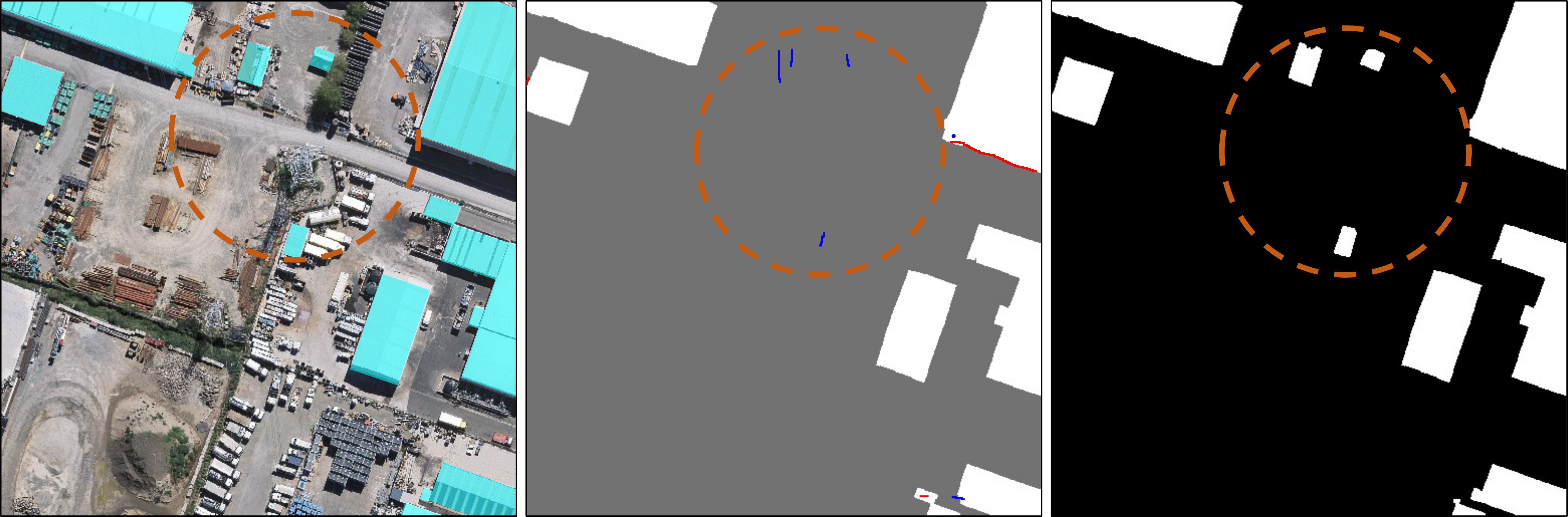}\caption{}
	\end{subfigure}	
	\begin{subfigure}[b]{0.8\linewidth}
		\includegraphics[width=\linewidth]{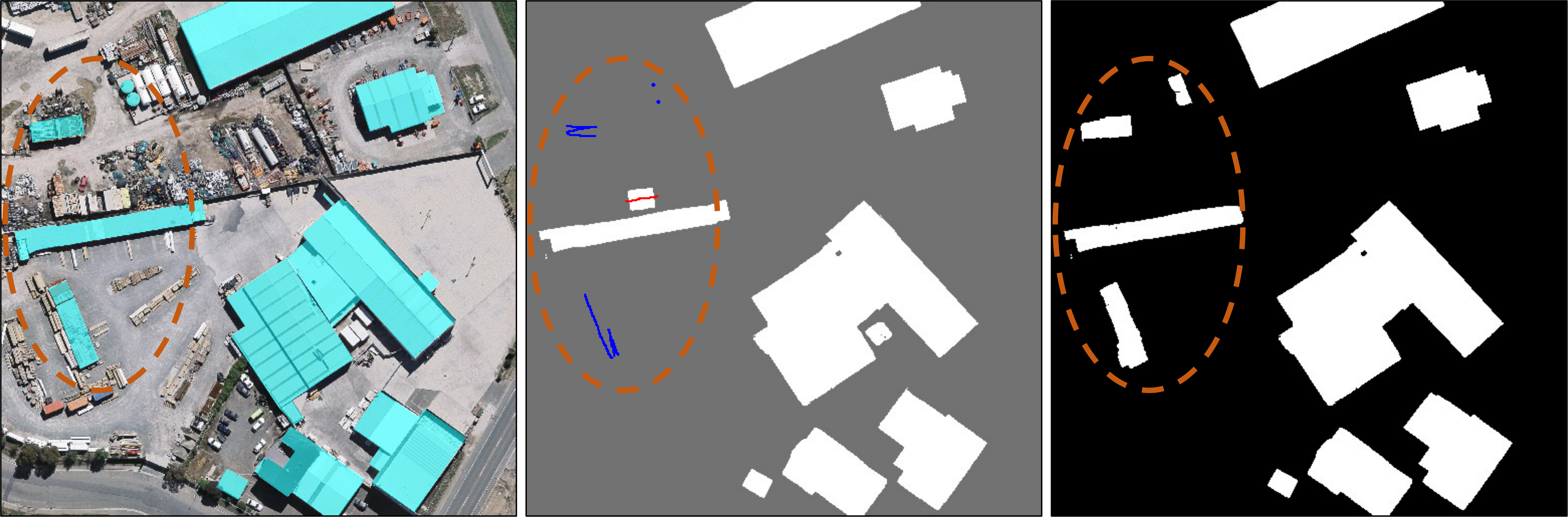}\caption{}
	\end{subfigure}	
	\begin{subfigure}[b]{0.8\linewidth}
		\includegraphics[width=\linewidth]{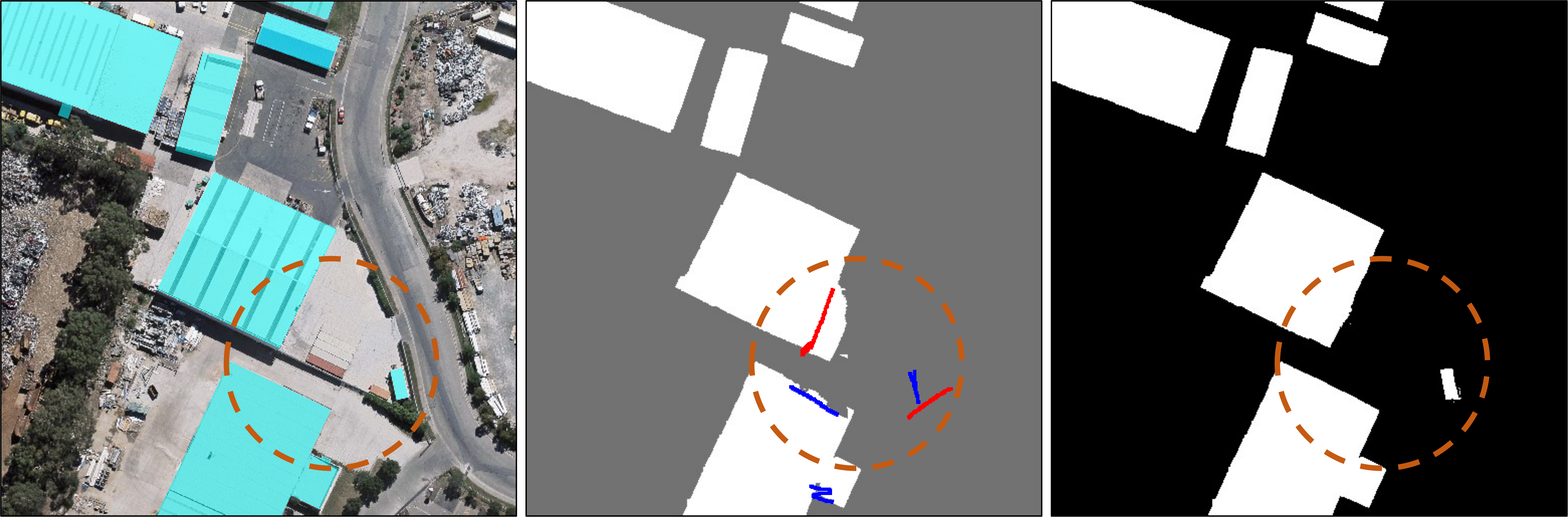}\caption{}
	\end{subfigure}	
	\begin{subfigure}[b]{0.8\linewidth}
		\includegraphics[width=\linewidth]{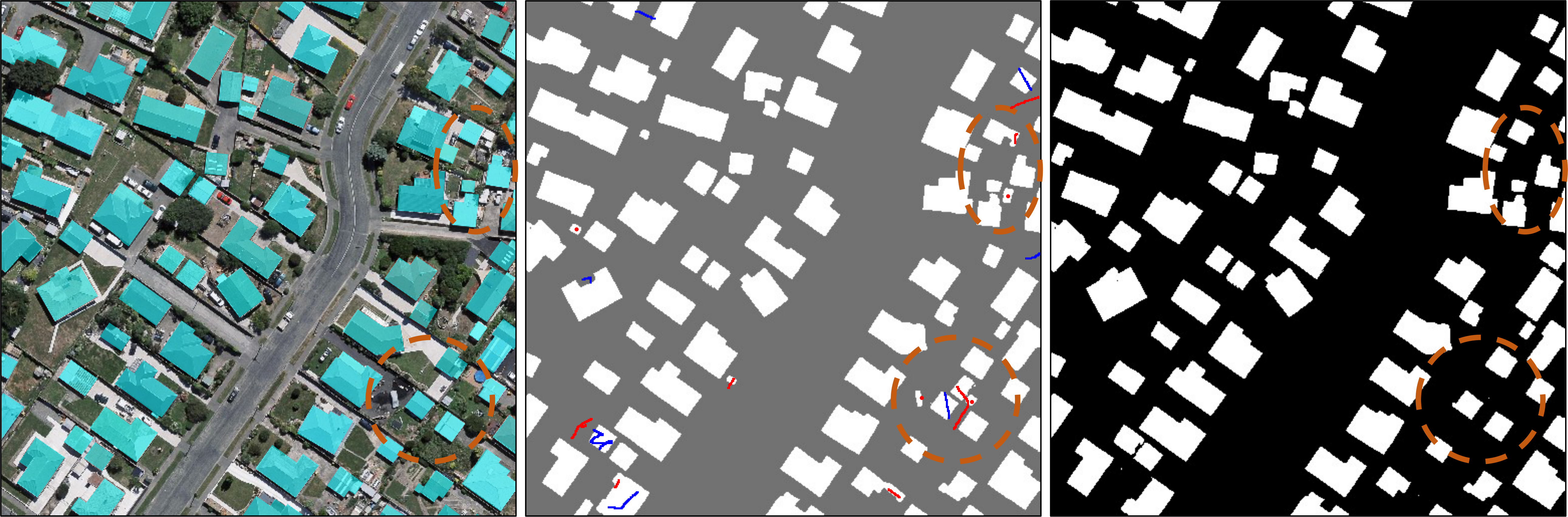}\caption{}
	\end{subfigure}	
	\caption{Examples of interactive optimization. From left to right: input images with ground truth footprints filled in cyan, initial results accompanied by interactive strokes, and the optimized results.}
	\label{fig:edit}
\end{figure}

Owing to the intricacy of roof types, environmental disruptions, occlusions, and noise, it is inevitable that segmentation outputs will be imperfect, which significantly hinders the adoption of current methods in real-world applications. In our work, we enable operators to edit the segmentation results using the auxiliary information generated through our approach\footnote{More interactive editing operations are available at \url{https://vrlab.org.cn/~hanhu/projects/spgraph/}}. We showcase several typical scenarios before and after interactive optimization in Fig. \ref{fig:edit}, and the qualitative results are also provided in Table \ref{tab:edit}. In practical implementations, we overlay the detected results on images with transparent shading. Operators draw strokes across the image, and when the mouse hovers over a pixel, the corresponding superpixel is highlighted. It is evident that substantial improvements can be achieved through minimal manual interaction, and various types of errors are rectified, including missed small roofs (Fig. \ref{fig:edit}a), false-positive detections (Fig. \ref{fig:edit}b), imperfect boundaries (Fig. \ref{fig:edit}c), and mixed errors in complex scenes with small buildings (Fig. \ref{fig:edit}d).

\begin{table}[H]
\centering
\caption{Quantitative improvements of the interactive optimizations corresponding to Fig. \ref{fig:edit}.}
\	\begin{tabular}{cclllll}
	\hline
	\multicolumn{1}{l}{} & \multicolumn{1}{l}{}          & \multicolumn{2}{c}{Pixel-level metrics} & \multicolumn{3}{c}{Vector-level metrics}     \\ \cline{3-7} 
	\multicolumn{1}{l}{} & \multicolumn{1}{l}{\#Strokes} & F1 score(\%)       & IoU(\%)            & AP50(\%)       & WC(\%)       & BF(\%)       \\ \hline
	\multirow{2}{*}{a}   & \multirow{2}{*}{9}            & 92.66              & 86.32              & 85.71          & 85.17        & 23.90        \\
						 &                               & 95.68(+3.02)       & 91.71(+5.39)       & 100.00(+14.29) & 91.70(+6.53) & 25.10(+1.20) \\ \hline
	\multirow{2}{*}{b}   & \multirow{2}{*}{7}            & 92.83              & 86.61              & 70.00          & 86.95        & 22.39        \\
						 &                               & 95.83(+3.00)       & 91.99(+5.38)       & 72.73(+2.73)   & 88.98(+2.03) & 23.60(+1.21) \\ \hline
	\multirow{2}{*}{c}   & \multirow{2}{*}{6}            & 96.18              & 92.64              & 85.71          & 92.52        & 25.80        \\
						 &                               & 98.20(+2.02)       & 96.47(+3.83)       & 100.00(+14.29) & 94.19(+1.67) & 26.94(+1.14) \\ \hline
	\multirow{2}{*}{d}   & \multirow{2}{*}{19}           & 93.30              & 87.43              & 84.71          & 88.86        & 25.12        \\
						 &                               & 95.24(+1.94)       & 90.89(+3.46)       & 97.34(+12.63)  & 90.76(+1.90) & 25.57(+0.45) \\ \hline 
	\end{tabular}
\label{tab:edit}
\end{table}

\section{Conclusion }
\label{s:con}
In this paper, we have presented a learning-based, semi-automatic building footprint detection algorithm that utilizes superpixels as the fundamental segmentation units, thereby ensuring improved preservation of building boundaries. By integrating both the superpixel segmentation and building semantic generation tasks within a single, multi-task network, our approach streamlines the process and enhances its efficiency. Furthermore, the superpixel graph construction within the network facilitates subsequent manual refinement. Our experimental findings reveal the effectiveness of the proposed method, showcasing its ability to produce more precise building outlines while also demonstrating significant potential for integration into practical applications. Despite its current reliance on a post-processing step for generating vectorized polygons, an intriguing avenue for future research involves the combination of the superpixel representation with an end-to-end vectorization module, which could potentially elevate the algorithm's performance and applicability.
\section*{Acknowledgments}
This work was supported in part by the National Natural Science Foundation of China (Project No. 42230102, 42071355, 41871291).

\bibliographystyle{model2-names}
\bibliography{main}

\end{document}